\documentclass[lettersize,journal]{IEEEtran}
\usepackage{amsmath,amsfonts}
\usepackage{algorithmic}
\usepackage{algorithm}
\usepackage{array}
\usepackage[caption=false,font=normalsize,labelfont=sf,textfont=sf]{subfig}
\usepackage{textcomp}
\usepackage{stfloats}
\usepackage{url}
\usepackage{verbatim}
\usepackage{graphicx}
\usepackage{cite}
\usepackage{widetext}
\usepackage{booktabs}
\usepackage{color, colortbl}
\usepackage{pifont}
\usepackage{bm}
\usepackage{multirow}
\usepackage{booktabs}
\usepackage{array}
\usepackage[font=small]{caption}

\begin{document}

\title{A Safe Self-evolution Algorithm for Autonomous Driving Based on Data-Driven Risk Quantification Model}

\author{Shuo Yang, Shizhen Li, Yanjun Huang$^*$, Hong Chen,~\IEEEmembership{Fellow,~IEEE}


\thanks{

Shuo Yang is with School of Automotive studies, Tongji University, Shanghai 201804, China (e-mail:shuo\_yang@tongji.edu.cn

Shizhen Li is with School of Automotive studies, Tongji University, Shanghai 201804, China (e-mail: 2231580@tongji.edu.cn)

Yanjun Huang is with School of Automotive studies, Tongji University, Shanghai 201804, China, and is also with Frontiers Science Center for Intelligent Autonomous Systems, Shanghai 200120, China (corresponding author: e-mail: yanjun\_huang@tongji.edu.cn)

Hong Chen is with the College of Electronics and Information Engineering, Tongji University, Shanghai 201804, China (e-mail: chenhong2019@tongji.edu.cn)

}


}



\maketitle

\begin{abstract}

Autonomous driving systems with self-evolution capabilities have the potential to independently evolve in complex and open environments, allowing to handle more unknown scenarios. However, as a result of the safety-performance trade-off mechanism of evolutionary algorithms, it is difficult to ensure safe exploration without sacrificing the improvement ability. This problem is especially prominent in dynamic traffic scenarios. Therefore, this paper proposes a safe self-evolution algorithm for autonomous driving based on data-driven risk quantification model. Specifically, a risk quantification model based on the attention mechanism is proposed by modeling the way humans perceive risks during driving, with the idea of achieving safety situation estimation of the surrounding environment through a data-driven approach. To prevent the impact of over-conservative safety guarding policies on the self-evolution capability of the algorithm, a safety-evolutionary decision-control integration algorithm with adjustable safety limits is proposed, and the proposed risk quantization model is integrated into it. Simulation and real-vehicle experiments results illustrate the effectiveness of the proposed method. The results show that the proposed algorithm can generate safe and reasonable actions in a variety of complex scenarios and guarantee safety without losing the evolutionary potential of learning-based autonomous driving systems.

\end{abstract}

\begin{IEEEkeywords}
Autonomous driving, Safety reinforcement learning, Risk quantification
\end{IEEEkeywords}

\section{Introduction}
\IEEEPARstart{A}{s} an important component of intelligent transportation systems (ITS), autonomous driving systems can complete driving tasks without human intervention through perception, decision-making and electronic control technologies \cite{teng2023motion}\cite{yang2021personalized}. Autonomous driving technology has the potential to enhance traffic efficiency, reduce the risk of traffic accidents, and offer sustainable solutions to transportation challenges, which has received widespread attention in recent years \cite{hu2023planning}\cite{li2023survey}. However, due to the challenges of algorithmic limitations and technological maturity, there are still many barriers to the realization of high-level autonomous driving. It is necessary to further research related technologies to ensure the safety and efficiency of autonomous driving systems in a variety of complex scenarios \cite{yurtsever2020survey}.

In recent years, self-evolution algorithms with experience storage and learning upgrade as the core ideas have attracted more and more attention. Reinforcement learning(RL), as an important part of self-evolution algorithms, has been widely used in many fields and has reached beyond human performance in some areas. It can be seen that the autonomous driving algorithms using reinforcement learning techniques have the potential to adapt to more complex scenarios \cite{cao2023continuous}\cite{feng2023dense}\cite{zhu2020safe}. However, due to the exploration-exploitation mechanism and the black box nature of reinforcement learning algorithm, it is difficult to guarantee the safety in the process of evolutionary learning and deployment application. This is absolutely unacceptable for a safety-critical system such as autonomous driving \cite{kiran2021deep}.

Safe reinforcement learning is a machine learning method that aims to enable a system to learn in uncertain or dangerous environments while minimizing the risk of producing undesirable outcomes \cite{gu2022review}\cite{garcia2015comprehensive}. The approach consists of two main categories, one of which is to consider optimization criteria that minimize the violation of constraints during the reinforcement learning process \cite{achiam2017constrained}\cite{yang2021wcsac}, and the other is to make stochastic exploration in the safety domain \cite{chen2022runtime}\cite{zhang2023evaluating}. Recently, many researches have applied safety reinforcement learning methods to autonomous driving tasks. These methods utilize a priori knowledge, either through historical data or human assistance to improve security during learning \cite{wang2019lane}\cite{brosowsky2021safe}\cite{aksjonov2023safety}. {Regarding the method of changing the optimization criterion, Yang et al. proposed a model-free security reinforcement learning method through Neural Barrier Certificate, which minimizes constraint violation in the process of policy collection through joint learning of policy and Neural Barrier Certificate \cite{yang2023model}. As for the design of the safety layer, the methods mainly include control barrier function methods \cite{cheng2019end}\cite{9478933}, rulebook methods \cite{hwang2022autonomous}\cite{gu2022constrained} and formalization methods \cite{leung2021using}. Zhang et al. proposed a safety reinforcement learning method for autonomous vehicles based on Barrier Lyapunov Function(BLF), which reasonably organized and incorporated BLF items into the optimized inverse control method, and constrained the state variables within the designed safety region during the learning process \cite{zhang2022barrier}. Zhang et al. proposed a safety checker based on a responsibility sensitive safety model that provides reinforcement learning agent with fallback safety actions in unsafe situations during the training and evaluation phases \cite{zhang2022safe}. Cao et al. proposed a decision-making framework called Trustworthy Improvement Reinforcement Learning (TiRL), which combines reinforcement learning and rule-based algorithms to allow self-improvement while maintaining better system performance \cite{cao2022trustworthy}.

However, although the above methods can consider safety in the algorithm learning process, the method of changing the optimization criterion can only reduce the constraint violation, and it is still difficult to ensure the process safety in the reinforcement learning algorithm, which cannot fully meet the high safety requirements of autonomous driving. At the same time, safety constraint methods may lead to insufficient exploration of the agent, which leads to difficulties in achieving optimal performance. The design of safety constraints based on manual definitions may also lead to conservative policies, resulting in performance loss \cite{ye2021meta}\cite{lv2022safe}. To solve this problem, a feasible solution is to design an adjustable safety limits, and design a safety limits adjustment criterion based on the risk quantitative value in order to alleviate the problem of balancing safety and performance that exists in evolutionary algorithms.

At present, there are two main methods for quantifying risk: rule-based methods and artificial potential field-based methods. The rule-based methods aim to evaluate the safety of autonomous driving systems in different situations by formulating a series of rules and indicators to quantify potential risks \cite{chen2019intelligent}\cite{dey2009desired}. Some definitions, such as Time-to-collision (TTC) \cite{nadimi2020evaluation}, Time-to-Lane-Change (TLC) \cite{wissing2017probabilistic}, Responsibly-sensitive Safety (RSS) \cite{koopman2019autonomous}, etc. have been proposed for calculating risk indicators. However, in some cases, these risk indicators can only support simple driving operations \cite{kilicarslan2018predict}. The method based on artificial potential field, on the other hand, combine logic rules and physical models to quantify risk scenarios by simulating potential risk factors through potential energy calculations \cite{wang2014concept}. However, the above methods still require more domain expertise and manual rule formulation, which limits their application in complex and dynamic traffic scenarios \cite{wang2021risk}. Hence, it makes sense to study risk quantification methods based on data-driven approaches that autonomously extract key features from data through learning methods, without the need for human predefinition of rules, and thus have potential in terms of real-world scenario adaptability \cite{chen2023follownet}.

To sum up, in this paper, a safety self-evolution algorithm for autonomous driving based on a data-driven risk quantification model is proposed. The method combines the risk quantization method and the safety self-evolution decision-control method in order to solve the problem of limited evolutionary performance of safety reinforcement learning methods in dynamic scenarios. An attention mechanism model based on human risk cognition is proposed to quantify the risk of the surrounding environment. In order to avoid the adverse impact of overly conservative safety limits on the self-evolutionary performance of autonomous driving algorithms, we introduce the concept of Safe Critical Acceleration (SCA). On the basis of this concept, a safety-evolution decision-control integration algorithm with adjustable safety limits is designed, and the proposed risk quantization model is integrated into it. The proposed method can generate safe and reasonable action policies in a variety of complex scenarios, while ensuring that the self-evolutionary potential of the learning autonomous driving system is not compromised.

The contributions of this study are summarized as follows:

1) This paper propose a risk quantification method based on a data-driven approach involves autonomously extracting key features from the data through machine learning methods, providing safety assessment indicators that consider human risk perception.

2) This paper propose a safety self-evolution method with adjustable safety limits integrating risk quantification indicators, which allows the algorithms to guarantee safety without sacrificing evolutionary performance.

3) This paper build a virtual-real interaction platfrom with the high-fidelity simulation software and the real autonomous vehicles, and use it to verify the effectiveness of the algorithm.

The paper is organized as follows. The proposed framework is introduced in Section II. The modeling of data-driven risk quantification model is introduced in Section III. The descriptions of safe self-evolution algorithm with adjustable safety limits are proposed in Section IV. In Section V, our method is verified and compared in simulation, and section VI concludes this paper.

\section{Proposed Framework}\label{sec:overall}

The proposed self-evolution algorithm for autonomous driving safety based on a data-driven risk quantification model aims to generate safe, efficient, and reasonable driving policies. The algorithm can not only ensure  safety during learning and deployment phase, but also effectively prevent the sacrifice of performance improvement ability. The input of the proposed algorithm is the information of the surrounding traffic environment, and the output is the safe control action. The main components are described in Fig. \ref{overall_architecture}, including a data-driven risk quantization model and a safety-evolution decision-control integration algorithm with adjustable safety limits.

The risk quantification model in this framework is designed as a transformer architecture. The proposed model simulates human risk perception by formulating a reinforcement learning problem that generates safety quantification indicators as output. Then, the output indicators are provided to the safety-evolution decision-control integration algorithm. The decision-control integration algorithm consists of three components, including an actor-critic framework, a local planner, and an adjustable safety limits mechanism. The actor-critic framework consists of an actor network and a critic network, which implement the algorithm evolution by maximizing the cumulative return. A local planner is combined with n adjustable safety limits mechanism to output reasonable actions that satisfy dynamic adjustable safety limits. The framework can reasonably and dynamically adjust the safety limit according to the surrounding traffic conditions, so that the performance improvement ability of the algorithm can be maintained as much as possible while ensuring the safety.

\begin{figure*}[ht!]
\centering
\begin{tabular}{c}
\includegraphics[width=0.87\textwidth]{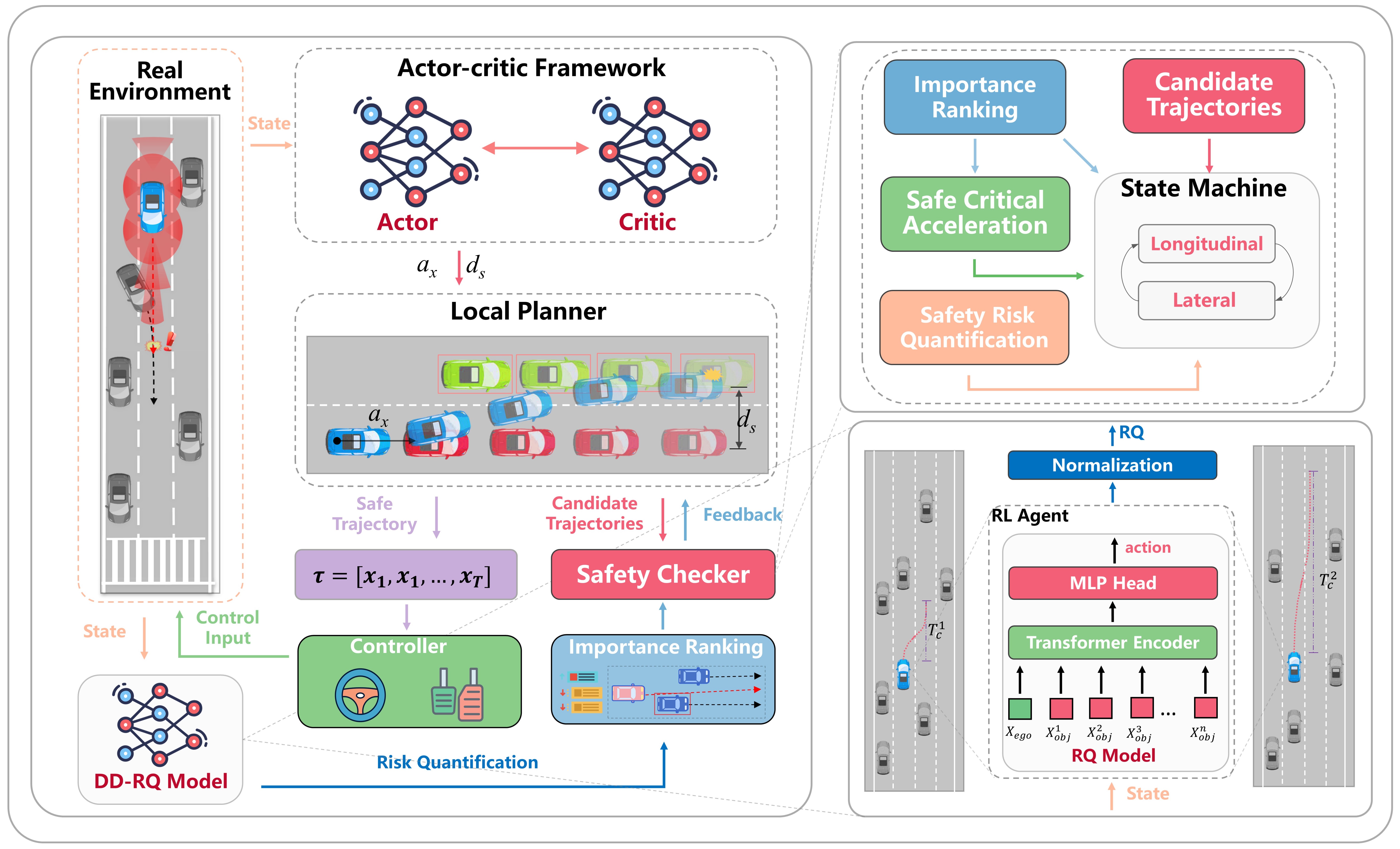}
\end{tabular}
\caption{Overall architecture of safe self-evolution algorithm for autonomous driving based on data-driven risk quantification model.}
\label{overall_architecture}
\end{figure*}

\section{Data-Driven  Risk Quantification Model}\label{sec:SQR_model}

\subsection{Problem Definition}

In this paper, we propose a risk quantification model modeling approach for assessing whether an autonomous vehicle is likely to enter an unsafe state during driving. With the goal of evaluating whether the current state is safe, the most intuitive idea is to consider the influence of the surrounding traffic situation on itself. However, it is worth noting that the risk quantification is not only related to the current state, but also to the driver's driving ability. An intuitive example is that in the same risk scenario, a skilled driver or higher-level autonomous driving system can be safe, but a novice or lower-level autonomous driving system may be at greater risk.

Therefore, the input of the risk quantification model should be the surrounding environmental information and the driving agent itself, and the output should be the risk quantification results. To sum up, the risk quantification model $f_{RQ}$ can be defined as:

\begin{equation}
\label{SQR_define}
\Lambda  = \left\{ {{f_{RQ}}\left( {s,A_\theta ^i} \right)|i = 1,2,...,n}\right\},\
\end{equation}
where $s$ is state, which is used to represent the surrounding traffic situation; $A_\theta $ is driving agent with $\theta$ as parameters. The state input $s$ and the driving agent $A_\theta $ together affect the risk quantification result $\Lambda$. $A_\theta $ is defined as:

\begin{equation}
{A_\theta } = \pi \left( s \right).\
\end{equation}

That is, the agent's policy output $\pi$ is related to the state input $s$.

\subsection{The mechanisms of human perception of risk}\label{subsec:RL agent}

Humans demonstrate high intelligence while driving, being able to subjectively estimate and evaluate risks. This ability allows humans to make effective decisions in complex and dynamic traffic environments. Therefore, by simulating human risk perception, the safety and responsiveness of autonomous driving systems can be enhanced.

By analyzing the above mechanism, the principle of constructing a risk quantification model can be deduced.

\begin{figure*}[ht!]
\centering
\begin{tabular}{c}
\includegraphics[width=0.8\textwidth]{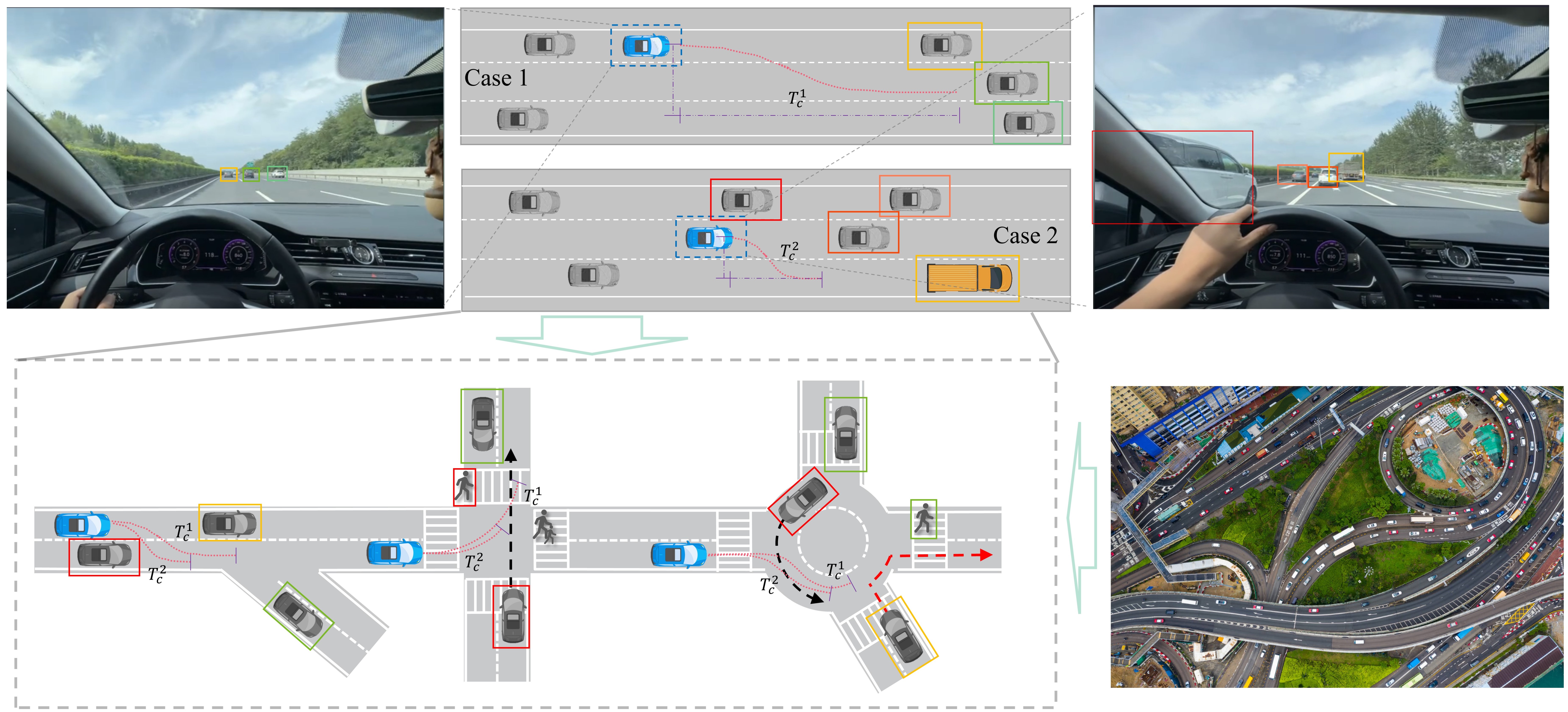}
\end{tabular}
\caption{The mechanisms of human risk perception.}
\label{The mechanisms of human risk perception}
\end{figure*}

\subsubsection{Human perception of risk}

Fig. \ref{The mechanisms of human risk perception} presents a schematic of the human risk perception mechanism. Taking the highway scenario as an example, the driver will complete various driving tasks such as following and overtaking by controlling the vehicle according to the movement state of the current traffic vehicle on the basis of judging the risk.

Case 1 illustrates a low risk situation. In this case, the driver focuses on the fact that the traffic ahead will have no effect on the ego vehicle due to the distance of the surrounding traffic. In terms of human cognitive behavior, the driver's attention is placed farther away, and therefore the trajectory planned in the driver's mind considers a more distant time span (denoted as $T_c^1$).

Case 2 illustrates a high risk situation. In this case, the driver focuses on the fact that the ego vehicle will be affected by the proximity of the surrounding traffic vehicles. In terms of human cognitive behavior, the driver's attention is placed near away, and therefore the trajectory planned in the driver's mind considers a closer time span (denoted as $T_c^2$).

Therefore, the construction of risk quantification model can be realized by simulating the human risk perception mechanism. Specifically, a human-like driving agent $\pi _\phi$ is established, which should have the following capabilities:

\begin{itemize}
  \item[\textbullet] Risk Quantification (RQ) : takes the current traffic situation as input and outputs a risk quantitative value given by the agent. The value is defined to range from 0 to 100\%, i.e., the larger the value is, the higher the risk is.
  \item[\textbullet] Importance Ranking (IR): All traffic participants within the agent perception range are ranked and scored according to the risk degree. The higher ranked participants are, the more likely they are to pose a risk to the ego vehicle, and therefore the more worthy of attention.
\end{itemize}

Considering the mechanism of human risk perception, we abstract the above problem and further propose the construction method of human-like agent based on this basis.

The above two cases are introduced with a straight road as an example. However, road conditions in urban driving scenarios are often complex, including congestion, off-ramps, intersections, roundabouts, and more. The Frenet coordinate system determines the position, direction, and speed on a road based on the tangent, normal, and tangent vectors at a point on a curve. This system is widely used in the decision-making and planning algorithms of autonomous driving due to its ability to correlate actual geometric features of the road and adapt to different road topologies \cite{werling2010optimal}. In summary, the proposed human risk perception mechanism and risk quantification method can be extended to more complex road environments based on straight roads through the Frenet coordinate system, as shown in Fig. \ref{The mechanisms of human risk perception}. In this paper, we focus on the design scheme of the proposed method on straight roads.

\subsubsection{Problem abstraction}

In low-risk situations, drivers' cognitive behavior tends to be more oriented towards long-term planning and smooth action strategies because safe driving is more easily achieved, whereas in high-risk situations, drivers may need shorter-term cognitive processes and more rapid reactivity, so drivers plan for shorter time spans and pay closer attention.

Hence, the above problem can be abstracted as: how to extract the smooth trajectory planned by the agent from the current position to complete the autonomous driving task, and how to use the corresponding time span of the trajectory to quantify the risk of the driver.

\begin{figure}[!t]
\centering
\includegraphics[width=3.5in]{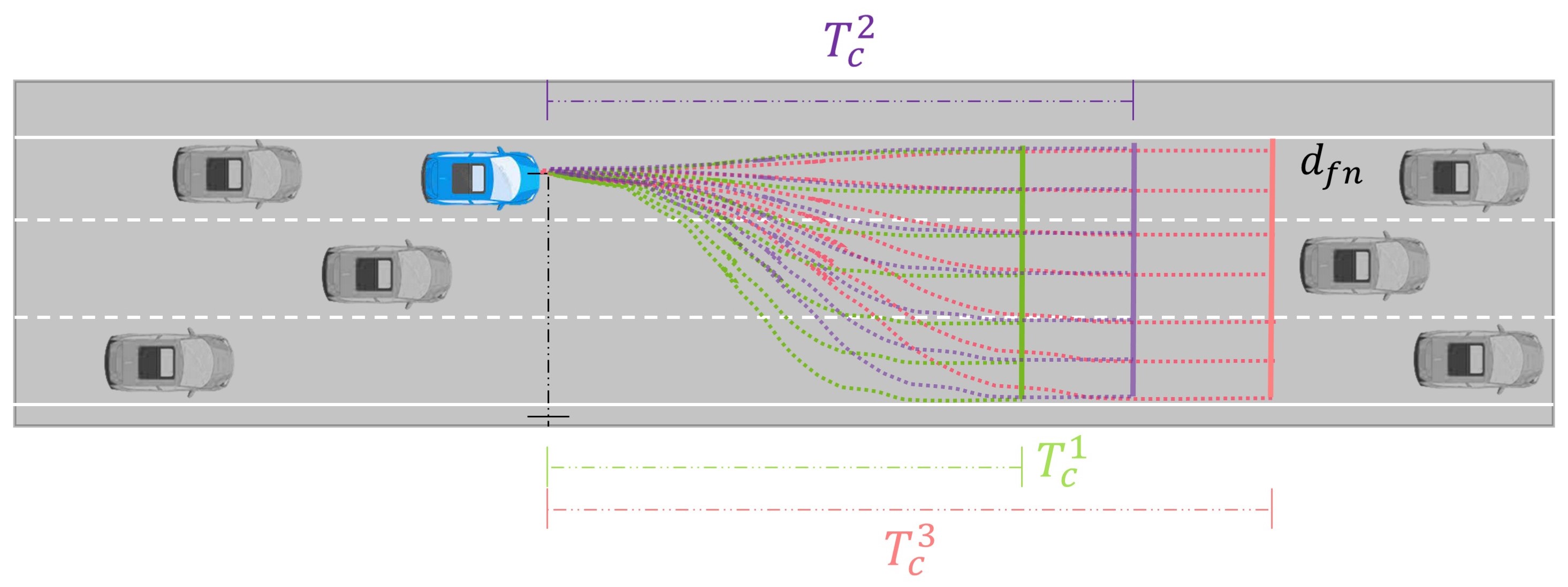}
\caption{Trajectory planning results for different planning times.}
\label{different_planning_times}
\end{figure}

The local trajectory planning process of the agent is shown in Fig. \ref{different_planning_times}. Quintic and quartic polynomial curves are respectively applied to describe the process of longitudinal and lateral trajectory planning, as shown in appendix.

The polynomial coefficients $p_d$ and $p_s$ can be solved by substituting the planning time $T_c$ with the boundary conditions in Eq. \ref{cal_polynomials 1} and Eq. \ref{cal_polynomials 2}. Fig. \ref{different_planning_times} illustrates the change of the longitudinal and lateral planning trajectories for different planning times ${T_c}^1$, ${T_c}^2$ and ${T_c}^3$. It can be seen that the smaller $T_c$ is, the shorter the planning time span is and the faster the response of the policy is.

As an effective self-learning method, RL has advantages in solving the decision and planning problem of autonomous driving, which can adapt to complex traffic environments and output efficient and safe driving policies. To sum up, the DD-RQ model proposed in this paper can be constructed as follows: constructing a human-like autonomous driving agent utilizing RL algorithm and then extracting planning time $T_c$ as a benchmark for risk quantification.

\subsection{Model construction}\label{subsec:RL problem design}

\subsubsection{Analysis of Cross-domain migration}

Limited by the learning efficiency issue and safety issue of reinforcement learning algorithms, it is difficult to collect enough data during actual operation and use it to train the DD-RQ Model. Therefore, training the RQ model in a simulation environment and deploying it in an application environment is a feasible solution.

However, considering the sim2real gap problem in both simulation and application environments \cite{kaspar2020sim2real}\cite{zhao2020sim}, the cross-domain migration of the RQ model needs to be analyzed. For the RQ model, its running domain is defined as $\Phi  = \varphi \left( {{P_{sc}},{G_{se}}} \right)$, where ${P_{sc}}$ is the scenario distribution and ${G_{se}}$ is the performance improvement ability of the algorithm.

It is clear that the ${P_{sc}}$ and ${G_{se}}$ of the self-evolution algorithm are similar in the simulation operational domain ${\Phi _s}$ and the application operational domain ${\mathord{\buildrel{\lower3pt\hbox{$\scriptscriptstyle\frown$}}
\over \Phi } _a}$. Therefore, the cross-domain migration process of the operational domain of the RQ model can be expressed as ${\Phi _s} \to {\mathord{\buildrel{\lower3pt\hbox{$\scriptscriptstyle\frown$}}
\over \Phi } _a}$ , and the following derivation can be made:

\textbf{\textit{Remark 1:}} when ${\Phi _s} \simeq {\mathord{\buildrel{\lower3pt\hbox{$\scriptscriptstyle\frown$}}
\over \Phi } _a}$ and ${\Phi _s} \to {\mathord{\buildrel{\lower3pt\hbox{$\scriptscriptstyle\frown$}}
\over \Phi } _a}$, the agent at the end of evolution have similar abilities, i.e., ${\mathord{\buildrel{\lower3pt\hbox{$\scriptscriptstyle\frown$}}
\over A} _\theta } \simeq \mathord{\buildrel{\lower3pt\hbox{$\scriptscriptstyle\frown$}}
\over A} $  .

Combined with the Eq. \ref{SQR_define}, it can be deduced that:

\textbf{\textit{Remark 2:}} when ${\mathord{\buildrel{\lower3pt\hbox{$\scriptscriptstyle\frown$}}
\over A} _\theta } \simeq \mathord{\buildrel{\lower3pt\hbox{$\scriptscriptstyle\frown$}}
\over A}$ , the risk quantitative value $\mathord{\buildrel{\lower3pt\hbox{$\scriptscriptstyle\frown$}}
\over \Lambda }  = {f_{RQ}}\left( {s,{{\mathord{\buildrel{\lower3pt\hbox{$\scriptscriptstyle\frown$}}
\over A} }_\theta }} \right) \simeq {f_{RQ}}\left( {s,{A_\theta }} \right) = \Lambda$, that is, the RQ model has the approximation principle.

The sim2real gap in the conventional sense reveals that the problem of mismatch in policy distribution occurs when the policies learned from the simulation environment are directly deployed to the application environment. In contrast, the safety limit algorithm proposed in this paper will take the current actual state as input and guarantee security through safe guard, which implies that the impact due to the quantitative variability of risk can be effectively eliminated.

\subsubsection{RL problem formulation}

The decision and planning problem for autonomous driving can be modeled as an MDP problem. A RL problem is formulated for constructing the human-like agent ${\pi _\phi }$ proposed in Section \ref{subsec:RL agent}. Typical RL problem consists of three parts: state space design, action design, and reward design.

The state design needs to consider the information of the ego vehicle, the traffic vehicles and the pedestrian (if need). The design of state space $s$ is shown in Table \ref{tab:State space}:

\begin{table}[!t]
\caption{State Space Design}\label{tab:State space}
\centering
\begin{tabular}{>{\centering\arraybackslash}m{0.6cm} >{\centering\arraybackslash}m{7.9cm}}
\toprule
State & Description \\
\midrule
$s_0$ & Normalized longitudinal distance of ego vehicle, ${s_e}/{s_{max}}$\\
$s_1$  & Normalized lateral distance of ego vehicle, ${d_e}/{L_{width}}$  \\
$s_2$  & Normalized longitudinal velocity of ego vehicle, ${v_s}/{v_{max}}$\\
$s_3$  & Lateral velocity of ego vehicle, $v_d$\\
$s_4$  & Heading angle of ego vehicle, ${\varphi _e}$ \\
$s_{5i}$ & Normalized longitudinal relative distance of ego vehicle and traffic vehicle $i$, $\left( {{s_o}^i - {s_e}} \right)/s_{max}^o$\\
$s_{5i+1}$ & Normalized lateral relative distance of ego vehicle and traffic vehicle $i$, $\left( {{d_o}^i - {d_e}} \right)/3 \cdot {L_{width}}$\\
$s_{5i+2}$  & Normalized longitudinal relative velocity of ego vehicle and traffic vehicle $i$, $\left( {v_s^i - {v_s}} \right)/v_{\max }^o$\\
$s_{5i+3}$ & Lateral relative velocity of ego vehicle and traffic vehicle $i$, $v_d^i - {v_d}$ \\
$s_{5i+4}$  & Heading angle of traffic vehicle $i$, $\varphi _o^i$\\
$s_{5i+3j+2}$ & Normalized longitudinal relative distance of ego vehicle and pedestrian\\
$s_{5i+3j+3}$ & Normalized lateral relative distance of ego vehicle and pedestrian \\
$s_{5i+3j+4}$ & Lateral velocity of pedestrian\\
\bottomrule
\end{tabular}
\end{table}

In this paper, $i = 1,2,3,4,5$. Considering the actual situation of the lane change scenario, the maximum distance of the ego vehicle is set to ${s_{max}} = 760m$, the maximum relative distance between the ego vehicle and the traffic vehicle is set to $s_{max}^o = 200m$, the maximum relative velocity between the ego vehicle and the traffic vehicle is set to $v_{\max }^o = 11.11m/s$.

The design of the action space needs to consider the goal of the autonomous driving task. The goal of trajectory planning for autonomous driving is to find a safe trajectory without collision. Combined with the analysis in Section \ref{subsec:RL agent}, the action space of RL problem is designed as follows:

\begin{equation}
\label{action_driving_agent}
\textnormal{a} = \left[ {\begin{array}{*{20}{c}}
{{T_c}}&{{d_{fn}}}&{{{\dot s}_f}}
\end{array}} \right],\
\end{equation}
where $T_c$ is planning time, ${d_{fn}}$ and ${\dot s_f}$ are the expected lateral offset and expected longitudinal velocity after time $T_c$, respectively. A simple PID controller is applied for tracking the desired trajectory generated by the RL algorithm.

The design of the reward function plays a crucial role in RL problems, which directly affects the performance of the algorithm. In order to effectively guide the learning of the agent, the formulation of the reward function should follow several key principles.

Firstly, positive or negative rewards that are too isolated may cause the algorithm to fall into unreasonable local optimal solutions, so both positive and negative rewards must be considered and the relationship between them must be balanced. Secondly, the design of the reward function should be consistent with the optimization goal of the agent, and appropriate rewards should be provided throughout the training process to avoid the problem of reward sparsity. In summary, the reward function is designed for the trajectory planning problem of autonomous driving.

The reward function is designed including the speed reward, the collision penalty and the lane deviation penalty, as detailed in the Appendix. The total reward function is: 

\begin{equation}
r = {r_s} + {r_c} + {r_{ld}}\
\end{equation}

\begin{figure}[!t]
\centering
\includegraphics[width=3.3in]{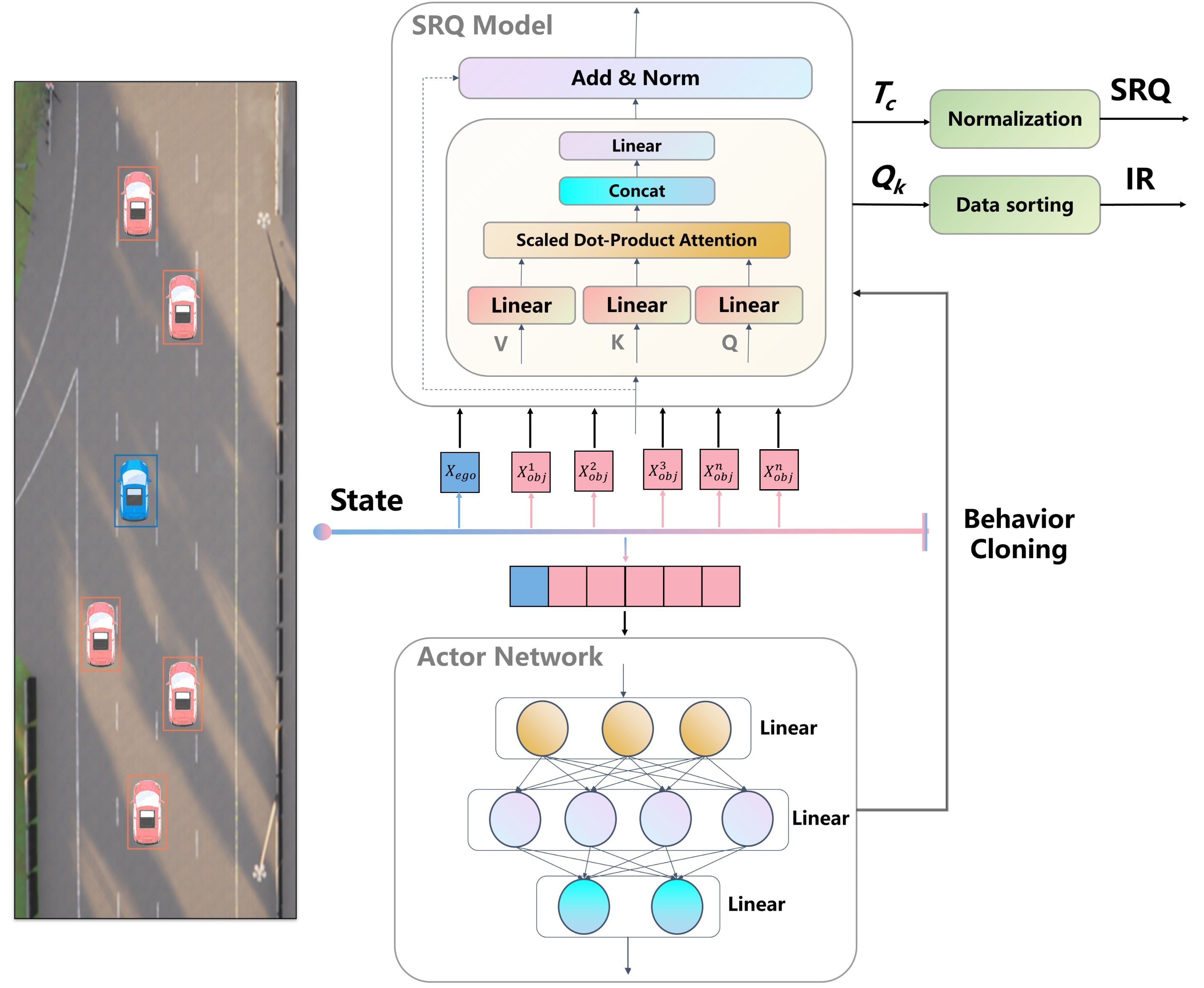}
\caption{Overall architecture of data-driven risk quantification model.}
\label{overall_SQR_model}
\end{figure}

\subsubsection{Risk Quantification Model}

The construction of the risk quantification(RQ) model is shown in Fig. \ref{overall_SQR_model}. First, based on the designed RL problem in this section, the soft actor critic algorithm is applied to train the driving agent. The training is performed in a high-fidelity simulation software, and the traffic vehicles are set to move according to a realistic traffic flow model. When the average return is stable and no longer grows, the training process ends, at which time we get a driving agent that can control ego vehicle to realize speed control and lane change to avoid obstacles.

To obtain RQ and IR results, RQ networks are introduced. The network uses the transformer decoding architecture. The actor network in the driving agent model is selected for collecting data, and the RQ network is trained by the behavior cloning method. Due to the introduction of attention mechanism, we can obtain the correlation degree between each state quantity, and then obtain the IR for the input of the state.

Recently, the transformer model has gained a lot of attention. The core idea of the transformer model is to capture long-distance dependencies and global information in the input sequence through the self-attention mechanism. By assigning different weights to each position in the sequence, the model can better capture important parts of the sequence. The attention mechanism not only helps to better understand the intrinsic structure of the input sequence, but also makes the model interpretable.

The proposed RQ model is constructed based on the transformer encoder model . The Transformer encoder  consists of alternating layers of  self-attention and MLP block. Layer norm (LN) is applied before every block, and residual connections after every block.

The input of the neural network is as described in the state space design in this section, including the relative position and relative speed information of ego vehicle as well as all traffic vehicles. The input vector can be expressed as follows:

\begin{equation}
{\rm X} = \left[ {{X_{ego}},X_{obj}^1,X_{obj}^2,...,X_{obj}^n} \right].\
\end{equation}

The standard transformer receives as input a 1D sequence of token embeddings. The embedding project $E$ is applied to extracted feature from input vector:

\begin{equation}
{h_0} = \left[ {{X_{ego}}E;X_{obj}^1E;X_{obj}^2E,...,X_{obj}^nE} \right],\begin{array}{*{20}{c}}
{}&{E \in {^{P \times D}}}
\end{array},\
\end{equation}
where ${X_{ego}}$ and $X_{obj}^i,i = 1,2,...,n$  are the state information of the ego vehicle and the traffic vehicles, respectively.

The transformer encoder model can be expressed as:

\begin{equation}
\label{MSA}
{\bf{z}}_\ell ^\prime  = {\mathop{\rm MSA}\nolimits} \left( {{\mathop{\rm LN}\nolimits} \left( {{{\bf{z}}_{\ell  - 1}}} \right)} \right) + {{\bf{z}}_{\ell  - 1}}\begin{array}{*{20}{c}}
{}&{\ell  = 1...L},
\end{array}\
\end{equation}

\begin{equation}
{{\bf{z}}_\ell } = MLP\left( {{\mathop{\rm LN}\nolimits} \left( {{\bf{z}}_\ell ^\prime } \right)} \right) + {\bf{z}}_\ell ^\prime \begin{array}{*{20}{c}}
{}&{\ell  = 1...L},
\end{array}\
\end{equation}

\begin{equation}
{\bf{y}} = {\rm{LN}}\left( {{\bf{z}}_L^0} \right).\
\end{equation}

In Eq. \ref{MSA}, the self-attention criterion divides the input embedding into three vectors $V$, $K$ and $Q$. The scaled dot-product attention is calculated as follows:

\begin{equation}
\label{QKV}
\Theta  = Attention(Q,K,V) = {\mathop{\rm softmax}\nolimits} \left( {\frac{{Q{K^T}}}{{\sqrt {{d_k}} }}} \right)V,\
\end{equation}
where $\Theta$ is scores matrix, $Q$ is a query vector, $K$ is a key vector, $V$ is a value vector, and ${d_k}$ is a normalization. The complete RQ model is shown in Fig. \ref{overall_SQR_model}.

\begin{algorithm}[H]
\caption{Training process of RQ model}\label{alg:alg1}
\begin{algorithmic}
\STATE \textbf{Initialize:} DA dataset $O = \emptyset$, RQ network $\varphi$ with weights ${\theta _{RQ}}$;
\STATE \textbf{Input:} state input $X$, converging actor network ${\pi _\phi }$, traffic model ${\tau _{tf}}$ with parameters $\sigma$;
\STATE \textbf{Set parameters:} Driving agent dataset size ${n_{DA}}$, max data collection episode number ${n_{eposide}}$, batch size ${n_b}$, epoch number ${n_{epoch}}$;
\STATE \textbf{for} \emph{i=1, 2, ...,}${n_{eposide}}$ \textbf{do}
\STATE \hspace{0.5cm}Random choose traffic model parameters $\sigma$
\STATE \hspace{0.5cm}Generate random traffic flow using ${\tau _{tf}}$ in simulation
\STATE \hspace{0.5cm}Using ${\pi _\phi }$ to collect $\left( {X,{\textnormal{a}^{[1]}}} \right)$
\STATE \hspace{1.0cm}$O \leftarrow O \cup \left\{ {\left( {X,T_c = {\textnormal{a}^{[1]}}} \right)} \right\}$
\STATE \hspace{1.0cm}\textbf{if} $len(O) > {n_{DA}}$ \textbf{then}
\STATE \hspace{1.0cm}break
\STATE \textbf{end for}
\STATE \textbf{for} \emph{i=1, 2, ...,}${n_{epoch}}$ \textbf{do}
\STATE \hspace{0.5cm}Random sample $B = {\left( {X,T_c = {\textnormal{a}^{[1]}}} \right)_{i = 0:{n_b} - 1}}$ from $O$
\STATE \hspace{0.5cm}Predict $\hat T_c = {\varphi _{{\theta _{RQ}}}}\left( X \right)$
\STATE \hspace{0.5cm}Compute MSE loss ${\cal L} = {\left\| {T_c - \hat T_c} \right\|^2}$
\STATE \hspace{0.5cm}Update ${\theta _{RQ}}$ by ${\hat \nabla _{{\theta _{RQ}}}}{{\cal L}_\varphi }({\theta _{RQ}})$ using Adams optimizer
\STATE \textbf{end for}
\end{algorithmic}
\label{alg1}
\end{algorithm}

The training process of the RQ model is shown in Algorithm \ref{alg:alg1}. Firstly, the driving agent dataset $O$ is collected. To ensure the model generalization, the random traffic flow is generated in the simulation environment by randomizing the parameters $\sigma$ of the traffic model ${\tau _{tf}}$. The converging actor network ${\pi _\phi }$ is deployed to the ego vehicle to perform the simulation, obtain the two-tuples $\left( {X,{\textnormal{a}^{[1]}}} \right)$ and collect them into $O$. The RQ model ${\varphi _{{\theta _{RQ}}}}$ is trained with data from $O$. In line 13, a batch $B$ is randomly sampled. In line 14 and line 15, the loss function is calculated as the mean squared error (MSE) between the predicted and true values, where $a$ and $\hat a$ represent the ground truth and predicted values, respectively. The model is trained using the Adams optimizer.

Through the normalization calculation, the RQ value can be obtained as follows.

\begin{equation}
RQ = \frac{{{\varphi _{{\theta _{RQ}}}}\left( X \right) - {T_c}_{\min }}}{{{T_c}_{\max } - {T_c}_{\min }}} \times 100\%, \
\end{equation}
where ${T_c}_{\max }$ and ${T_c}_{\min }$ are the maximum and minimum values of ${T_c}$.

The scores matrix $\Theta$ can be calculated using Eq. \ref{QKV}

\begin{equation}
\label{scores_matrix}
\Theta  = \left[ {\begin{array}{*{20}{c}}
{\mu _{ego}^{1 \times 1}}&{\mu _{obj}^{1 \times 2}}& \cdots &{\mu _{obj}^{1 \times (n + 1)}}\\
{\mu _{obj}^{2 \times 1}}&{\mu _{att}^{2 \times 2}}& \cdots &{\mu _{at}^{2 \times (n + 1)}}\\
 \vdots & \vdots & \ddots & \vdots \\
{\mu _{obj}^{n + 1 \times 1}}&{\mu _{at}^{n + 1 \times 2}}& \cdots &{\mu _{at}^{n + 1 \times (n + 1)}}
\end{array}} \right]\
\end{equation}

As can be seen from Eq. \ref{scores_matrix}, where $\mu $ is the matrix of $5 \times 5$, and the lower right corner marker is used to distinguish the source of $\mu$. $\Theta$ can be used to characterize the feature correlation degree between each state. Since the correlation between the states of each vehicle is of most interest, the elements of the matrix $\mu$ are summed to obtain the new matrix ${\tilde \Theta ^{(n + 1) \times (n + 1)}}$.

To obtain the ranking of the feature correlation degree between ego and other traffic vehicles, the dimension $2$ to $n+1$ of the first row of ${\tilde \Theta ^{(n + 1) \times (n + 1)}}$ is taken to form a new vector, and the index of the ordering of vector elements is obtained. As shown in Eq. \ref{sort_result}:

\begin{equation}
\label{sort_result}
\Lambda  = argsort({\tilde \Theta ^{(n + 1) \times (n + 1)}}\left[ {2:end} \right], )
\end{equation}
where the index matrix $\Lambda$ is the vector of $1 \times n$, which is the IR output by the RQ model.

\section{Safe Self-Evolution Algorithm with Adjustable Safety Limit}

In this section, a safety-evolution decision-control integration algorithm with adjustable safety limits is proposed. The influence of overly conservative safety guard policy on the self-evolution ability is prevented by integrating the proposed RQ model.

\subsection{Agent implementation}

A RL problem based on soft actor critic algorithm is constructed to achieve integrated autonomous driving decision and control with evolutionary capability. It is worth emphasizing that there are some significant differences between the agent ${\pi _\omega}$ presented in this section and the driving agent ${\pi _\phi }$ described in section \ref{subsec:RL agent}. Specifically, the agent ${\pi _\omega}$ is used to deploy on the autonomous vehicle to actually control the ego vehicle to realize the driving task, and the driving agent ${\pi _\phi }$ is trained and deployed in the simulation to provide training data for the DD-RQ model. Hence, although these two agents share the framework of RL, they still have some differences in problem formulation and model design due to their different application goals.

\subsubsection{State space and action design}

For an autonomous driving system, the input information accepted by the algorithm is fixed, that is, the required information during the driving process is predetermined. Thus, the state space $\mathord{\buildrel{\lower3pt\hbox{$\scriptscriptstyle\frown$}}\over s}$ of ${\pi _\omega}$ is designed to be consistent with ${\pi _\phi }$, as shown in Table \ref{tab:State space}.

The design of action space needs to consider the actual goal of autonomous driving trajectory planning. For the agent ${\pi _\omega}$, different from Eq. \ref{action_driving_agent}, we directly choose the longitudinal acceleration of the ego vehicle $a_x$ as the control input. The new action space is designed as follows:

\begin{equation}
\mathord{\buildrel{\lower3pt\hbox{$\scriptscriptstyle\frown$}}
\over a}  = \left[ {\begin{array}{*{20}{c}}
{{d_{fn}}}&{{a_x}}
\end{array}} \right],\
\end{equation}
where ${d_{fn}}$ is the desired lateral offset and $a_x$ is the longitudinal acceleration of the ego vehicle. A simple PID controller is applied again to track the desired trajectory generated by the RL algorithm \cite{farag2020complex}.

\subsubsection{Reward design}

The reward function design is carried out on the basis of section \ref{subsec:RL problem design} and the Appendix. In the actual autonomous driving process, in addition to the speed reward $r_s$, collision penalty $r_c$ and lane departure penalty $r_{ld}$, it is also necessary to consider the driving risk penalty ${r_{risk}}$ and comfort reward $r_{cf}$.

As described in Section \ref{sec:overall}, the RQ model and the safety guard policy work together to ensure the safety. On this basis, if the collision risk is recognized, the driving risk penalty ${r_{risk}}$ will be triggered to take effect. As shown in Eq. \ref{risk_penalty}.

\begin{equation}
\label{risk_penalty}
{r_{risk}} = \left\{ {\begin{array}{*{20}{c}}
0&{\begin{array}{*{20}{c}}
{with}&{risk}
\end{array}}\\
{{\rho _{risk}}}&{risk}
\end{array}}, \right.\
\end{equation}
where ${\rho _{risk}}$ is driving risk penalty factor

The comfort reward ${r_{cf}}$ is designed to reduce the amplitude of the longitudinal and lateral actions as much as possible. To make the training easier, a normalization method is applied as shown in Eq. \ref{safety_reward}.

\begin{equation}
\label{safety_reward}
{r_{cf}} = {\rho _{cf}} \times \left[ {\frac{{{\mathop{\rm abs}\nolimits} \left( {d_{fn}^i - df_{fn}^{i - 1}} \right)}}{{2 \times {L_{width}}}} + \frac{{{\mathop{\rm abs}\nolimits} \left( {{a_x}_{\rm{ }}^i - {a_x}_{\rm{ }}^{i - 1}} \right)}}{{\bar a}}} \right],\
\end{equation}
where $d_{fn}^i$ and $d_{fn}^{i - 1}$ are the expected lateral displacements at the current time and the last time, ${a_x}_{\rm{ }}^i$ and ${a_x}_{\rm{ }}^{i - 1}$ are the longitudinal accelerations of the ego vehicle at the current time and the last time, ${L_{width}}$ is the lane width, $\bar a$ is the normalized coefficient of the longitudinal acceleration, and ${\rho _{cf}}$ is the comfort reward coefficient.

To sum up, the total reward $\mathord{\buildrel{\lower3pt\hbox{$\scriptscriptstyle\frown$}}\over r}$ is:

\begin{equation}
\mathord{\buildrel{\lower3pt\hbox{$\scriptscriptstyle\frown$}}
\over r}  = {r_s} + {r_c} + {r_{ld}} + {r_{risk}} + {r_{cf}}.\
\end{equation}

All reward coefficients s for ${\pi _\omega }$ are listed in Table III.

\begin{table}[!t]
\caption{Reward Coefficients Settings for ${\pi _\omega }$\label{tab:Reward Coefficients Settings_2}}
\centering
\begin{tabular}{cc}
\toprule
Reward coefficients terms & Symbol \& Value \\
\midrule
Driving risk  & ${\rho _{{risk}}} = -5.0$ \\
Comfort  & ${\rho _{cf}} = -0.5$ \\
\bottomrule
\end{tabular}
\end{table}

\subsection{Safe Critical Acceleration}

Considering that the rule-based safety limit logic will affect the learning effect of the algorithm, this paper proposes the concept of safe critical acceleration(SCA), which is used to integrate the DD-RQ proposed in Section III to realize the dynamic adjustment of safety limits.

The safety state of an autonomous vehicle is defined as the longitudinal distance between the ego vehicle and all traffic vehicles is greater than the safe distance ${S_{safe}}$. Combining the experience of experts and laws and regulations, the safe distance ${S_{safe}}$ is defined as:

\begin{equation}
{S_{safe}} = v_s^{ego} \times 3.6,\
\end{equation}
where $v_s^{ego}$ is the longitudinal speed.

\begin{figure}[!t]
\centering
\includegraphics[width=3.5in]{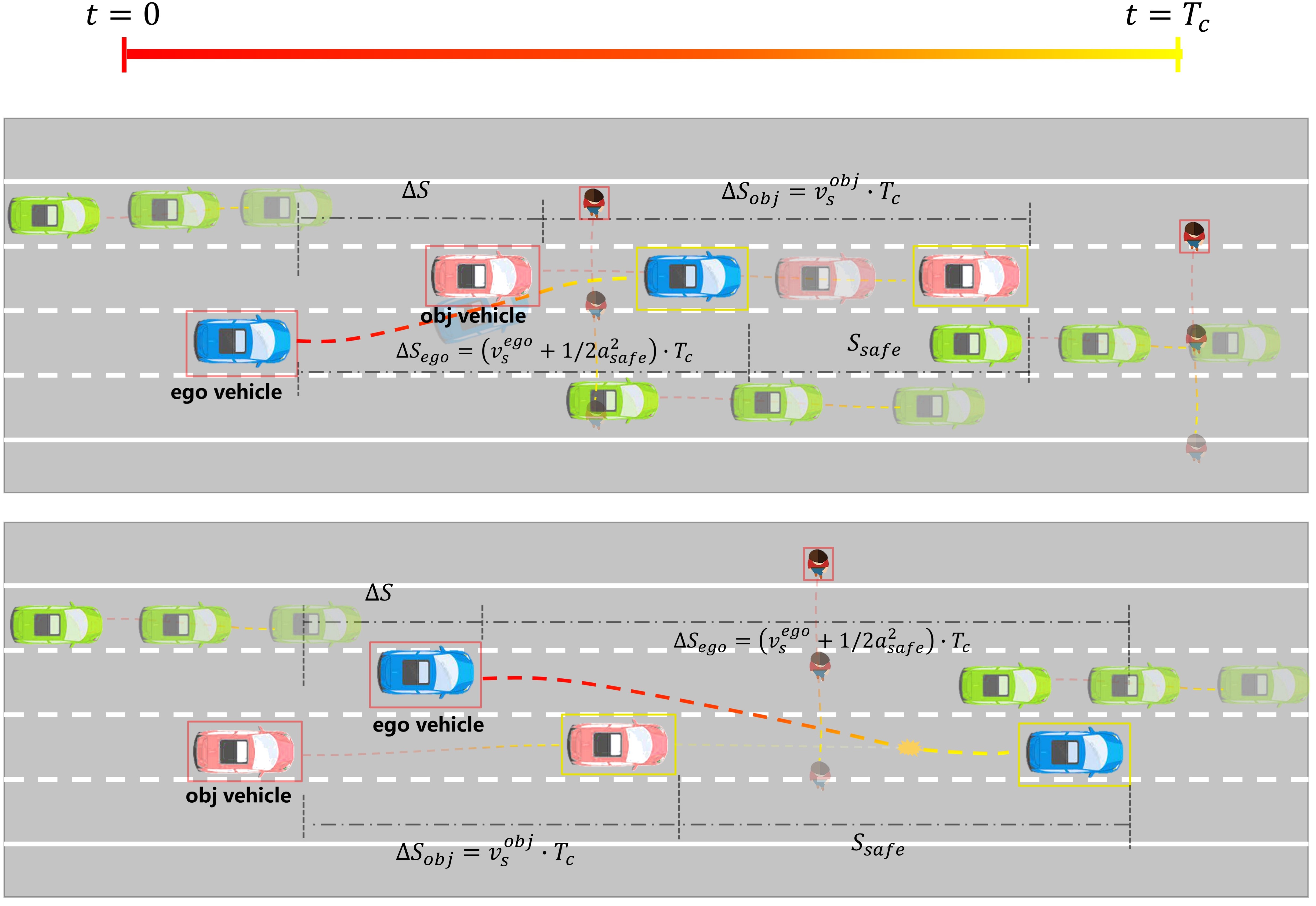}
\caption{Safe distance adjustment. When $\Delta S > 0$, the ego vehicle should slow down to adjust the safe distance, and when $\Delta S < 0$, the ego vehicle should accelerate to adjust the safe distance}
\label{Safe_distance_adjustment}
\end{figure}

So far, the safety evolution problem of the learning-based autonomous driving algorithm can be defined as follows: during the learning process of the algorithm, when the vehicle is considered to be in or about to be in an unsafe state, the system should reach back to the safe state through lateral or longitudinal control inputs after an appropriate adjustment time $T_c$. As shown in Fig. \ref{Safe_distance_adjustment}, it's the schematic diagram of the safety distance adjustment process. This process can be expressed as follows:

\begin{equation}
\label{safe_dis_adj}
\left\{ \begin{array}{l}
{S_{obj}} - {S_{ego}} + \Delta {S_{obj}} = \Delta {S_{ego}} + {S_{safe}},{S_{obj}} - {S_{ego}} \ge 0\\
{S_{ego}} - {S_{obj}} + \Delta {S_{ego}} = \Delta {S_{obj}} + {S_{safe}},{S_{obj}} - {S_{ego}} < 0
\end{array}, \right.\
\end{equation}
where ${S_{obj}}$ and ${S_{ego}}$ are the longitudinal displacements of the traffic vehicles and the ego  vehicle, $\Delta {S_{obj}}$ and $\Delta {S_{ego}}$ are the relative longitudinal displacements of the traffic vehicles and the ego vehicle in time $T_c$, respectively.

The safe distance adjustment process is regarded as a constant acceleration movement, and the acceleration in the adjustment process is defined as the safe critical acceleration ${a_{safe}}$. ${a_{safe}}$ is used to characterize the acceleration limit of the ego vehicle when it does not exceed the safe distance(that is, not leave the safe state). On the basis of Eq. \ref{safe_dis_adj}, the safe critical acceleration of the ego vehicle for the $i$-th traffic vehicles can be expressed as follows:

\begin{equation}
\label{SCA}
{a_{saf{e_i}}} = \frac{{2\left( {\Delta S - \frac{{\Delta S}}{{\left| {\Delta S} \right|}} \cdot {S_{safe}} + {T_c} \cdot \Delta v} \right)}}{{T_c^2}},\
\end{equation}
where $\Delta S = {S_{obj}} - {S_{ego}}$, $\Delta v = v_s^{obj} - v_s^{ego}$.

In Eq. \ref{SCA}, $T_C$ will affect the $a_{safe}$. Specifically:

\textcircled{1}: When $\Delta S > 0$, the ego vehicle should slow down to adjust the distance between the ego vehicle and the  preceding traffic vehicle to the safe distance ${S_{safe}}$. At this time, shortening the adjustment time $T_c$ will require the ego vehicle to pursue a safe distance ${S_{safe}}$ from the front vehicle by slowing down for a shorter time, so the required deceleration will be larger. At this point, the control output by the RL algorithm much less likely to satisfy the demand, so the safety guard is more likely to intervene.

\textcircled{2}: When $\Delta S < 0$, the ego vehicle should accelerate to adjust the distance between the ego vehicle and the following traffic vehicle to the safe distance ${S_{safe}}$. At this time, shortening the adjustment time $T_c$ will require the ego vehicle to pursue a safe distance ${S_{safe}}$ from the rear vehicle by accelerate for a shorter time, so the required acceleration will be larger. At this point, the control output by the RL algorithm much less likely to satisfy the demand, so the safety guard is more likely to intervene.

In summary, by adjusting ${T_c}$, ${a_{safe}}$ can be adjusted, which indirectly realizes the dynamic adjustment of the safety limits of the proposed evolutionary algorithm.

\begin{algorithm}[H]
\caption{The design of safety guard policy}\label{alg:alg2}
\begin{algorithmic}
\STATE \textbf{Initialize:} Intervention Flag $\Gamma  = False$, environment vehicles set ${D_{obj}} = \left[ {{v_1},{v_2},...,{v_n}} \right]$, safety-critical acceleration set of the preceding vehicles $O_{acc}^{pre} = \emptyset$, safety-critical acceleration set of the  following vehicles $O_{acc}^{follow} = \emptyset$;
\STATE \textbf{Input:} RQ model ${\varphi _{{\theta _{RQ}}}}$, state input $X$, RL action ${\rm{a = }}\left[ {{d_{fn}},{a_x}} \right]$;
\STATE \textbf{Set parameters:} Maximum acceleration $a_{max}$, minimum acceleration $a_{min}$;
\STATE Calculate ${f_{path}}({\rm{a}})$ using Eq. \ref{cal_polynomials 1} and Eq. \ref{cal_polynomials 2};
\STATE \textbf{for} \emph{$v_1$=1, 2, ...,}${D_{obj}}$ \textbf{do}
\STATE \hspace{0.5cm}$RQ,I{R_i} \leftarrow {\varphi _{{\theta _{RQ}}}}\left( X \right)$
\STATE \hspace{0.5cm}${T_c} = Linear(RQ)$
\STATE \hspace{0.5cm}Take $T_C$ as input and calculate ${a_{saf{e_i}}}$ using Eq. \ref{SCA}
\STATE \hspace{0.5cm}\textbf{if} $\Delta S > 0$ \textbf{do} $O_{acc}^{pre} \leftarrow O_{acc}^{pre} \cup \left\{ {\left( {{v_i},I{R_i},{a_{saf{e_i}}}} \right)} \right\}$
\STATE \hspace{0.5cm}\textbf{else} $O_{acc}^{follow} \leftarrow O_{acc}^{follow} \cup \left\{ {\left( {{v_i},I{R_i},{a_{saf{e_i}}}} \right)} \right\}$

\STATE \textbf{end for}
\STATE Find $a_{safe}^{pre}$ using $\max \left( {{a_{saf{e_i}}} + Linear(I{R_i})} \right)$ in $O_{acc}^{pre}$
\STATE Find $a_{safe}^{follow}$ using $\max \left( {{a_{saf{e_i}}} + Linear(I{R_i})} \right)$ in $O_{acc}^{follow}$
\STATE \textbf{if} $a_{safe}^{pre} < a_{safe}^{fllow}$ or $a_{safe}^{pre} < {a_{min}}$ or $O_{acc}^{follow} > {a_{max}}$ \textbf{do}
\STATE \hspace{0.5cm}$\Gamma  = True$
\STATE \hspace{0.5cm}${{\rm{a}}^{[0]}} = {d_{fn}} = 0$
\STATE \hspace{0.5cm}Calculate $f_{path}^{new}({\rm{a}})$ using Eq. \ref{cal_polynomials 1} and Eq. \ref{cal_polynomials 2}
\STATE \textbf{end if}
\STATE Repeat calculate $a_{safe}^{pre}$ and $a_{safe}^{fllow}$ using Eq. \ref{SCA}.
\STATE \textbf{if not} $a_{safe}^{follow} < {{\rm{a}}^{[1]}} = {a_x} < a_{safe}^{pre}$ \textbf{do}
\STATE \hspace{0.5cm}$\Gamma  = True$
\STATE \hspace{0.5cm}${{\rm{a}}^{[1]}} = {a_x} = a_{safe}^{pre}$
\STATE \textbf{end if}
\end{algorithmic}
\label{alg1}
\end{algorithm}

\subsection{Adjustable safety limits based on quantitative indicator}

\begin{figure*}[t!]
\centering
\subfloat[Dynamic dense traffic scenario]{\includegraphics[width=6.5in]{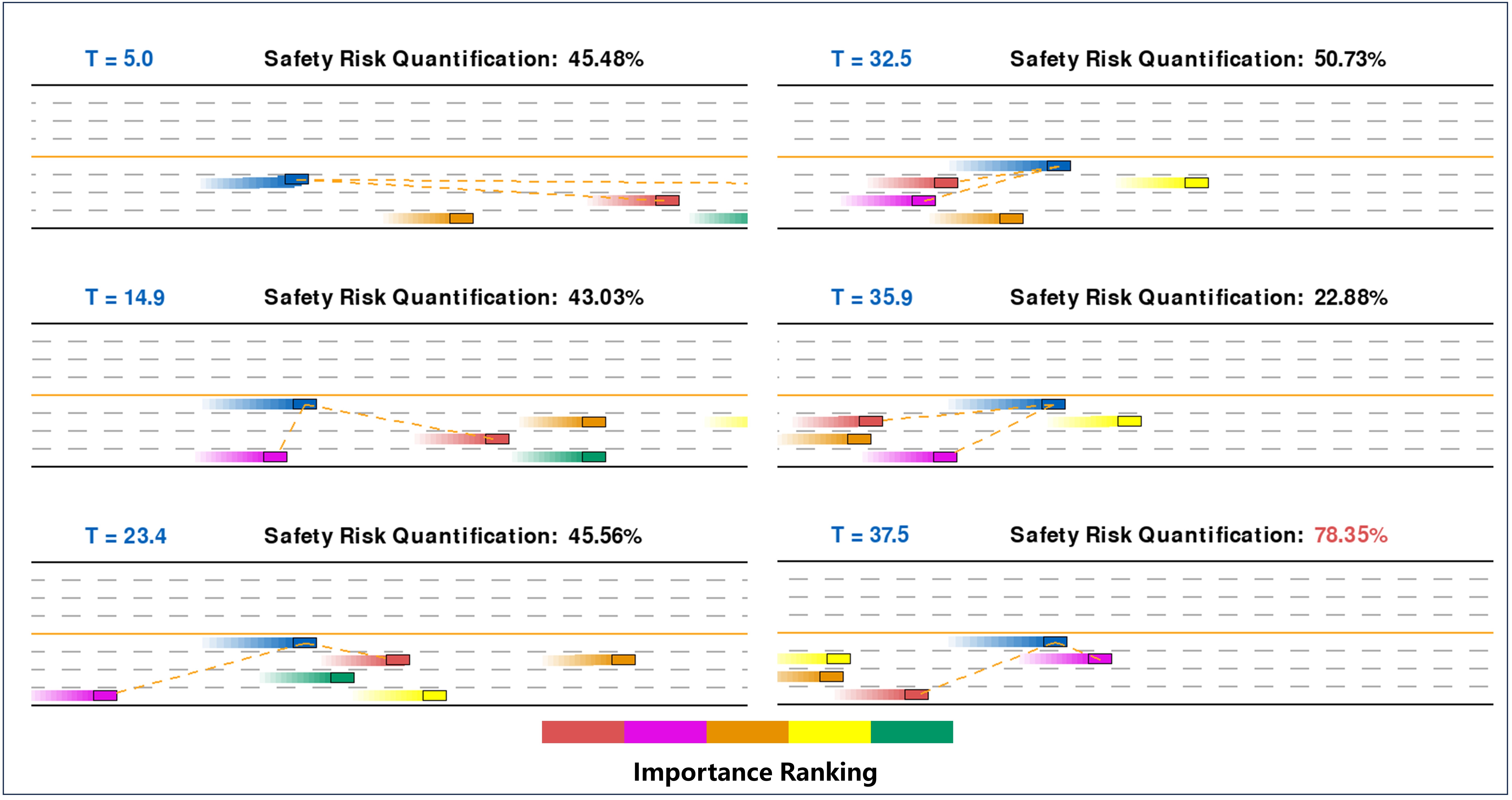}}
\label{dynamic dense traffic scenario}
\hspace{0.1in}
\subfloat[Mixed traffic scenario including pedestrian crossing]{\includegraphics[width=6.5in]{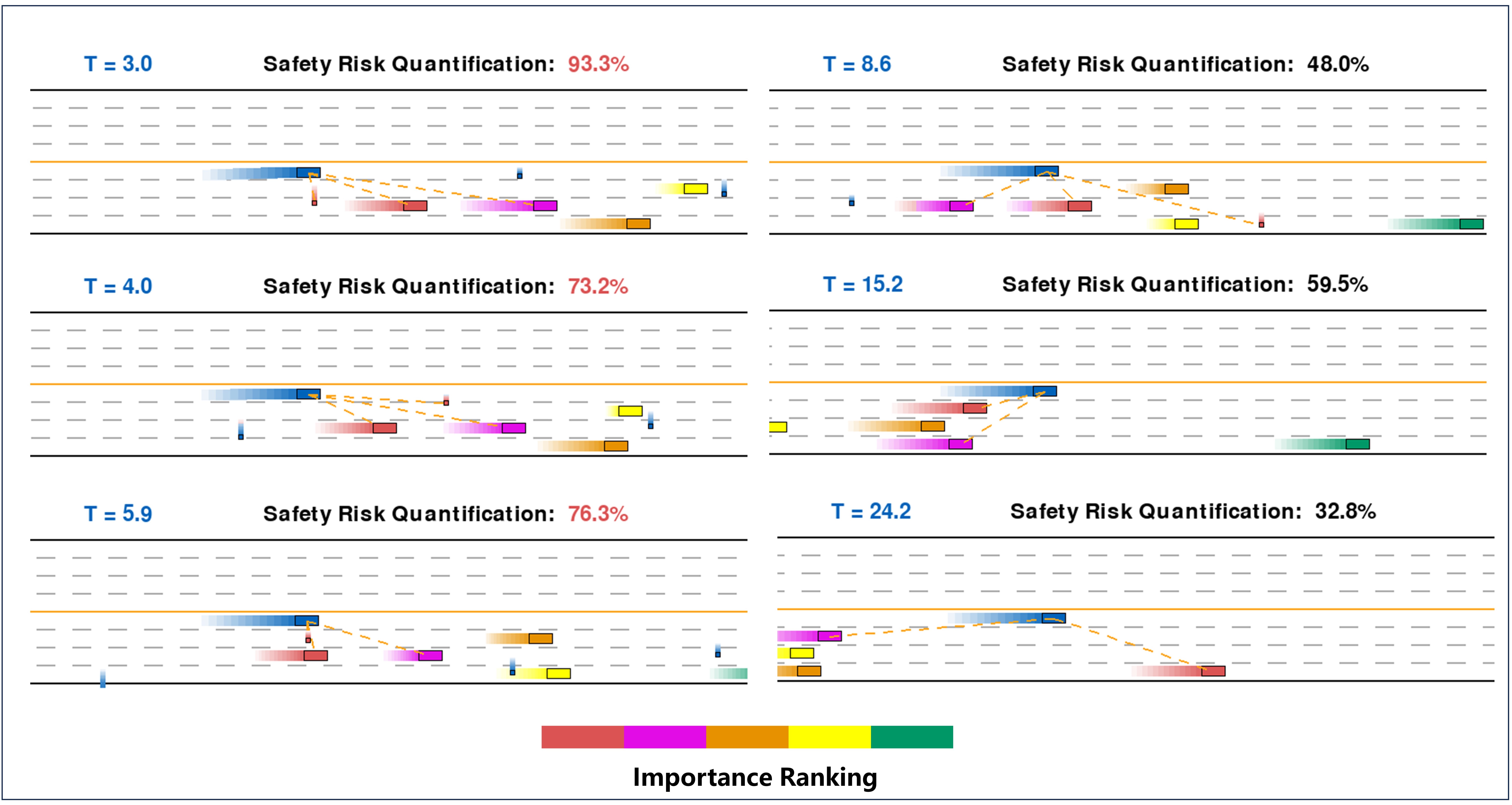}}
\label{mixed traffic scenario including pedestrian crossing}
\hspace{0.1in}
\caption{Visualization results of the DD-RQ model. (a) Dynamic dense traffic scenario. (b) Mixed traffic scenario including pedestrian crossing.}
\label{Visualization_results_of_the_DD-RQ}
\end{figure*}

\begin{figure}[!t]
\centering
\includegraphics[width=3.4in]{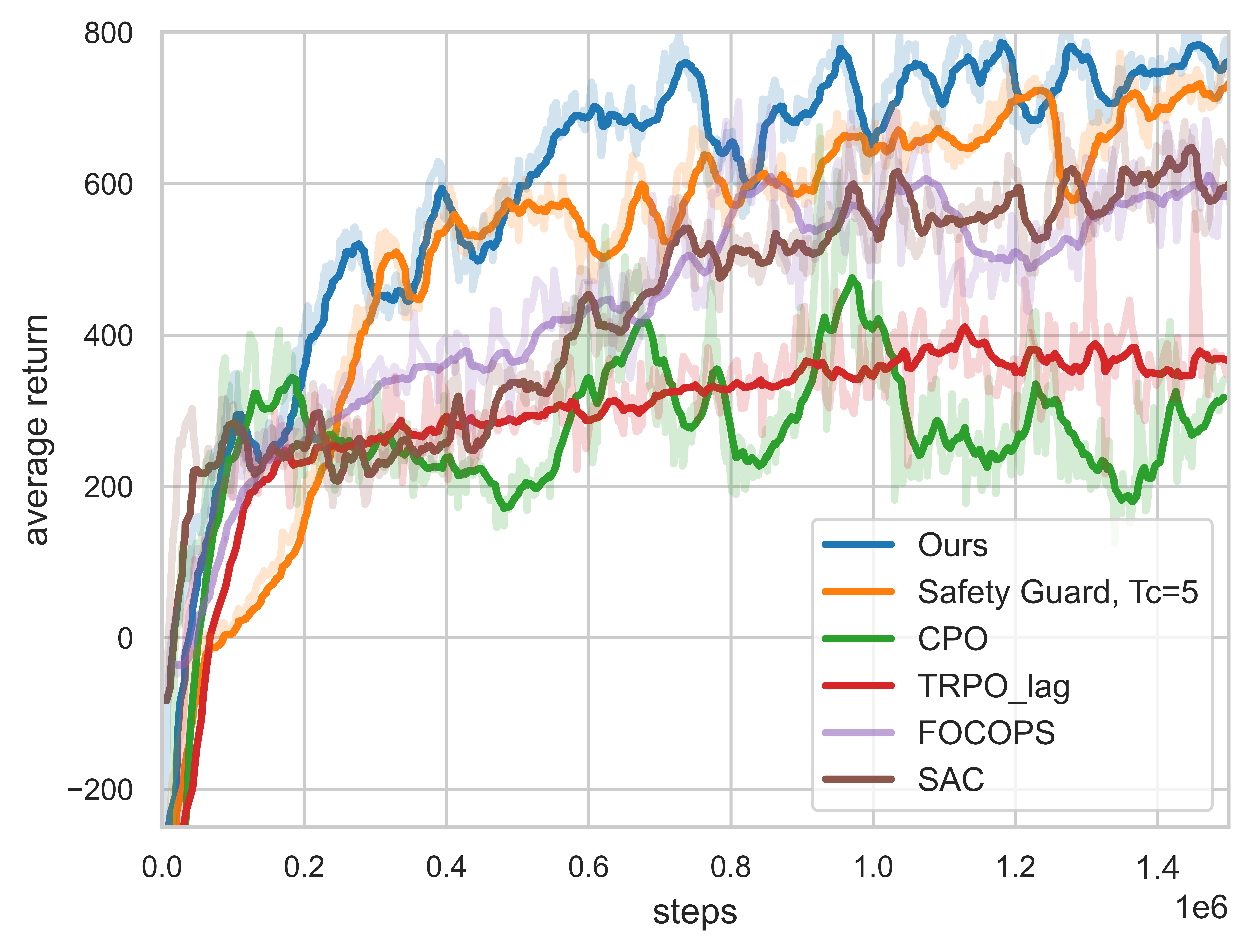}
\caption{Average return as a function of environment steps.}
\label{average return}
\end{figure}

\begin{figure*}[ht!]
\centering
\begin{tabular}{c}
\includegraphics[width=0.97\textwidth]{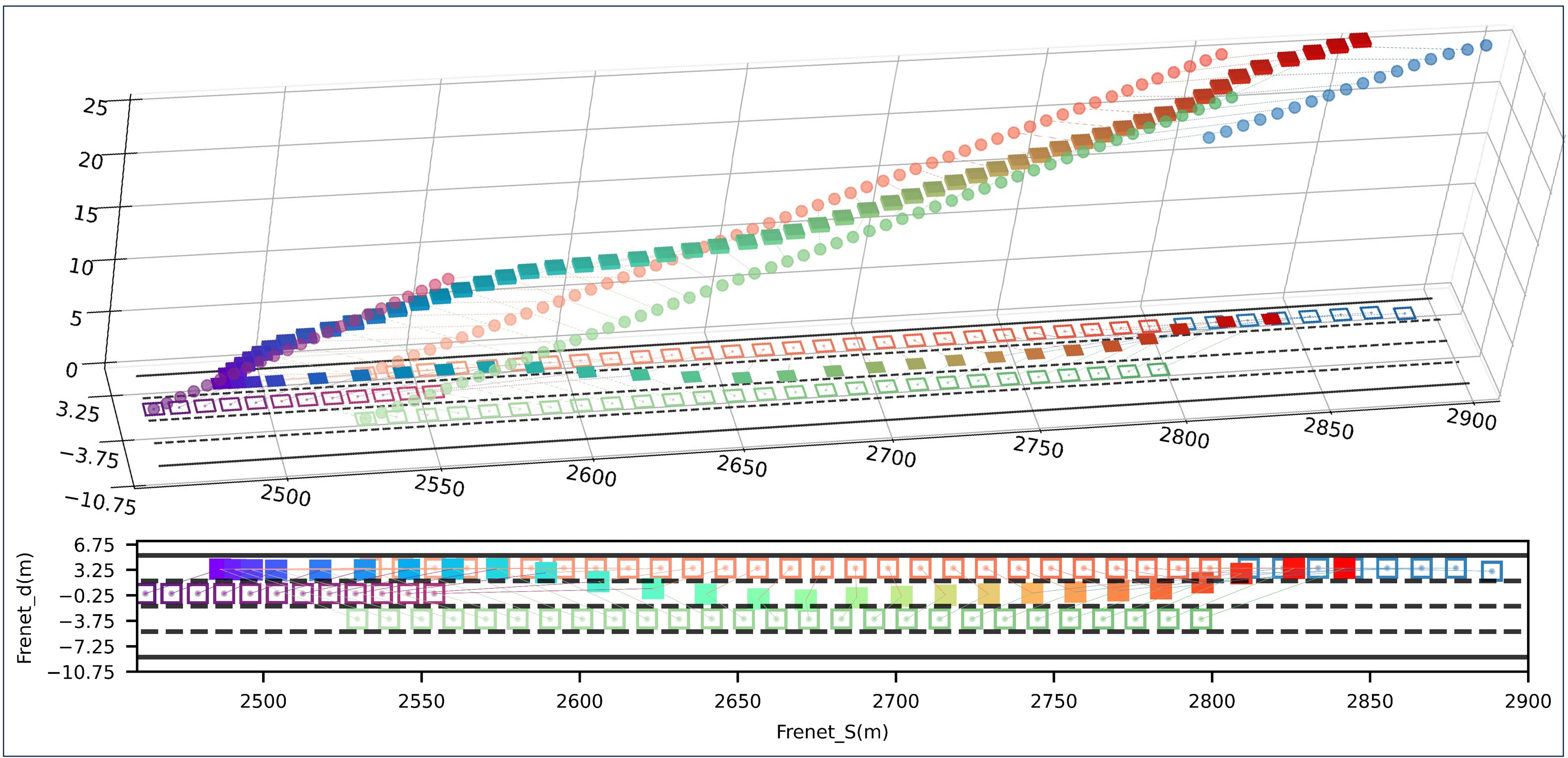}
\end{tabular}
\caption{Spatio-temporal diagram of ego vehicle and traffic vehicles trajectories in scenario a)}
\label{Spatio-temporal diagram_sa}
\end{figure*}

An excessively conservative safety guard intervention policy may have adverse effects on the self-evolution performance of autonomous driving algorithms. Therefore, combined with the RQ model proposed in Section \ref{sec:SQR_model}, this paper designs a safety self-evolution method with adjustable safety limits integrating risk quantification indicators

Specifically, the RQ model outputs a quantitative value of risk $RQ$ for the current traffic environment. $T_C$ is defined as a linear function of $RQ$ and is used to dynamically adjust the safety limits. In high risk scenarios, the safety limits are tightened. This is to ensure that the system is able to adopt a more conservative policy in high-risk scenarios, thus reducing the risk of potential accidents. Whereas in low risk scenarios, the safety limits are relaxed. This allows the system to explore more flexibly in a safe environment, thus enhancing its ability to self-evolve and learn. This adjustment  mechanism has the advantage of being clear and explainable to maximize the evolutionary potential of the algorithm while ensuring safety.

The proposed safety guard policy consists of two parts. Firstly, according to the initial action given by the RL algorithm, combined with the safe critical acceleration ${a_{saf{e_i}}}$, to determine whether there is a risk in the lateral action and needs to be intervened. Afterwards, if the lateral action is judged to be risky, it is set to maintain the current lane and further judge whether the longitudinal action is risky enough to intervene.

The intervention principle of the safety guard is shown in Algorithm \ref{alg:alg2}. The algorithm inputs include the RQ model obtained in Section \ref{sec:SQR_model}, the state input $X$ representing the surrounding environment, and the actions ${\rm{a = }}\left[ {{d_{fn}},{a_x}} \right]$ given by the RL algorithm. The main steps of the algorithm are as follows:

Line 1- Line 3: Initializes the algorithm and sets parameters.

Line 4- Line 10: Traverse all the traffic vehicles, calculate RQ, $I{R_i}$ and ${a_{saf{e_i}}}$ of each traffic vehicle. According to the relative position relationship between the traffic vehicles and the ego vehicle, the three-tuples $\left( {{v_i},I{R_i},{a_{saf{e_i}}}} \right)$ is stored into safety-critical acceleration set of the  following vehicles $O_{acc}^{follow}$ and safety-critical acceleration set of the preceding vehicles $O_{acc}^{pre}$ respectively. Line 6 shows the linearization process used to get the mapping from $T_c$ to RQ, where the parameters are $k_1$ and $b_1$.

Line 11- Line 12: Select the maximum ${a_{safe}}$ from $O_{acc}^{pre}$ and $O_{acc}^{follow}$. The selection index is ${a_{saf{e_i}}} + Linear(I{R_i})$, where the linearized parameters are $k_2$ and $b_2$.

Line 13- Line 17: Determine whether the RL output is risky at this time. If  true, the safety guard needs to intervene, then the lateral action will be set to keep driving in the current lane, i.e. ${{\rm{a}}^{[0]}} = {d_{fn}} = 0$, and recalculate $f_{path}^{new}({\rm{a}})$ using Eq. \ref{cal_polynomials 1} and Eq. \ref{cal_polynomials 2}

Line 18- Line 22: Continue to determine whether the RL's longitudinal action output is risky at this time. If  true, the safety guard needs to continue to intervene, at which point the longitudinal action is set to safety critical acceleration, i.e. ${{\rm{a}}^{[1]}} = {a_x} = a_{safe}^{pre}$.

\section{Experiments}

\subsection{Experiment setting}

In this section, the proposed safe self-evolution algorithm for autonomous driving based on data-driven risk quantification model is validated in a challenging three-lane stochastic traffic scenario. There are two evaluation scenarios, including a) a dynamic dense traffic scenario and b) a mixed traffic scenario including pedestrian crossing. In scenario a), the training scenario is set within a 180 $m$ range in front of the ego vehicle, with randomly generated traffic flows in the speed range 8 $m/s$ to 12 $m/s$. Scenario (b) is then based on scenario a), which generates pedestrians crossing the road according to random locations and random speeds, and makes them interact with the traffic flow. The training environment was built in the simulation software CARLA \cite{moghadam2020end}. The parameters of the algorithm are set as shown in Appendix. The video link is: https://www.youtube.com/watch?v=Q6bNhhWySO4

\begin{figure*}[ht!]
\centering
\begin{tabular}{c}
\includegraphics[width=0.97\textwidth]{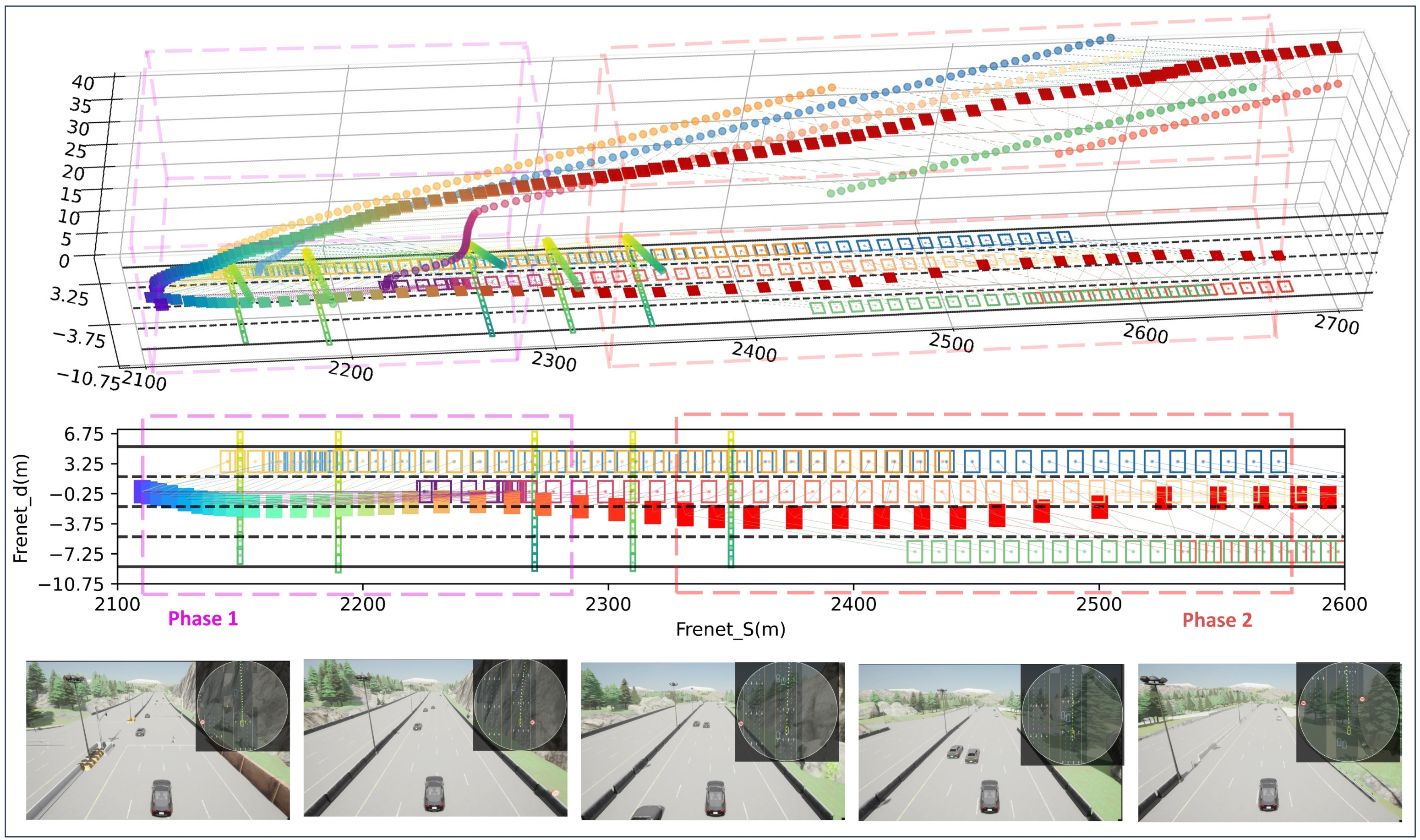}
\end{tabular}
\caption{Spatio-temporal diagram of ego vehicle and traffic vehicles trajectories in scenario b).}
\label{Spatio-temporal diagram_sb}
\end{figure*}

\begin{figure*}[ht!]
\centering
\begin{tabular}{c}
\includegraphics[width=0.95\textwidth]{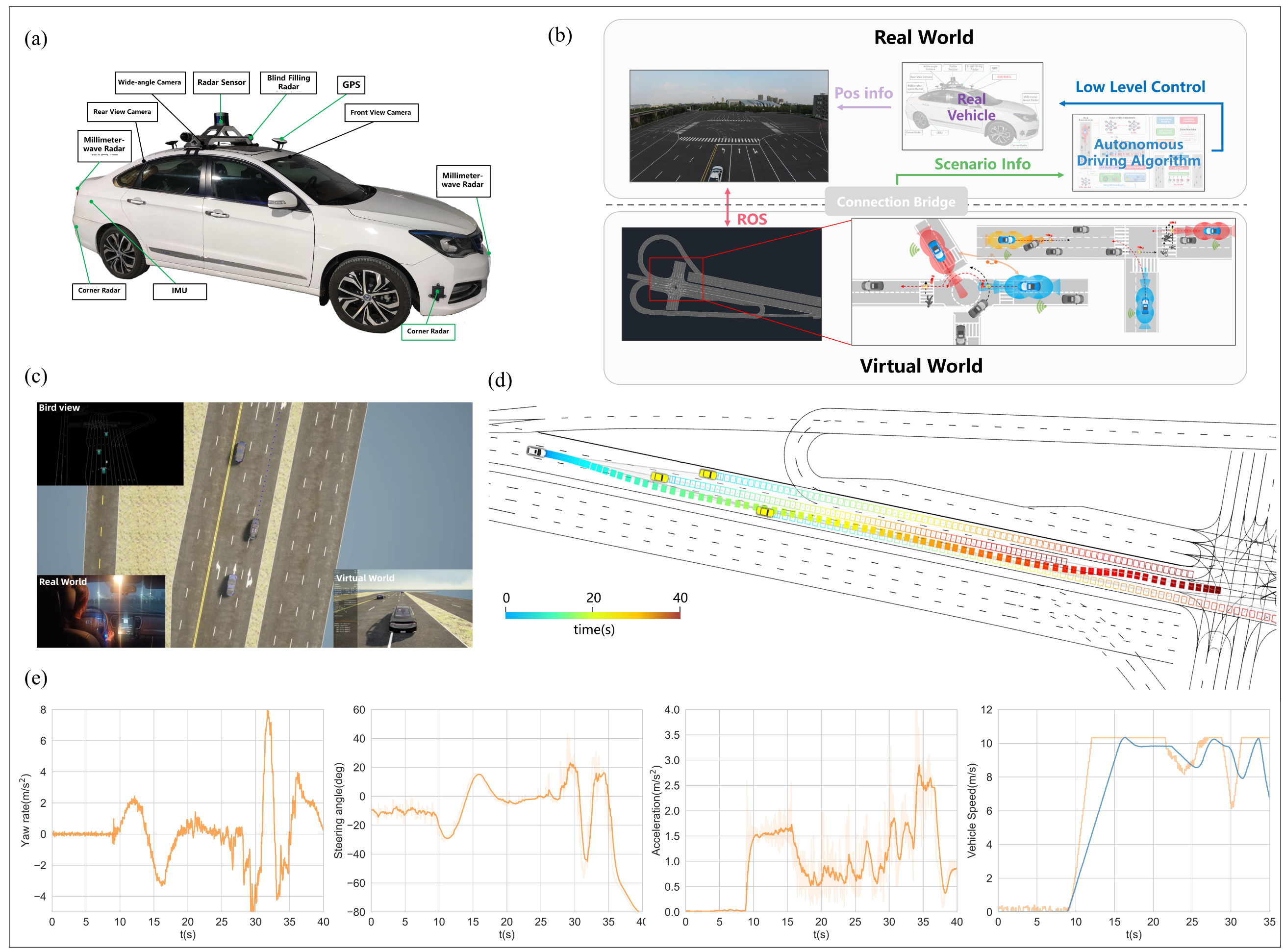}
\end{tabular}
\caption{Real vehicle test results. (a) The sensor configuration of the autonomous driving vehicle. (b) The overall architecture of the virtual-real interaction platform. (c) Experiment results in the real and virtual worlds. (d) Visualization of the trajectories of the vehicle and traffic vehicles in real vehicle test. (e) Sensor recording results, Yaw rate vs Timesteps, Steering angle vs Timesteps, Acceleration vs Timesteps, Vehicle speed vs Timesteps, respectively.}
\label{Real_vehicle_test_Results}
\end{figure*}

\begin{table*}[ht!]
\centering
\caption{Results of the ablation experiments}\label{tab:Results of the ablation experiments}
\begin{tabular}{>{\centering\arraybackslash}p{1.8cm}>{\centering\arraybackslash}m{1.3cm}>{\centering\arraybackslash}m{1.6cm}
>{\centering\arraybackslash}m{1.4cm}>{\centering\arraybackslash}m{1.5cm}>{\centering\arraybackslash}m{1.4cm}>{\centering\arraybackslash}m{1.4cm}
>{\centering\arraybackslash}m{1.4cm}}
\toprule
\multirow{2}{2.4cm}{\centering Stats} & \multirow{2}{*}{Stage} & \multirow{2}{*}{\textbf{Ours}} & \multicolumn{4}{c}{Safety Guard} &\multirow{2}{*}{ SAC} \\
\cmidrule(lr){4-7}
 & & & $T_c=1.0$ & $T_c=3.0$ & $T_c=5.0$ & $T_c=7.0$ & \\
\midrule
\multirow{2}{2.4cm}{\centering Collision Rate} & T & $\textbf{0\%}$ & $0\%$ & $2\%$ & $3\%$ & $0\%$ & $49.7\%$ \\
 & A & $\textbf{0\%}$ & $0\%$ & $0\%$ & $0\%$ & $0\%$ & $27.5\%$ \\
\midrule
\multirow{2}{2.4cm}{\centering Avg. Speed} & T &\bm{${12.58}$} \bm{$m/s$} & $11.80$ $m/s$ & $11.83$ $m/s$ & $11.61$ $m/s$ & $11.83$ $m/s$ & $12.80$ $m/s$ \\
 & A &\bm{${12.66}$} \bm{$m/s$} & $11.97$ $m/s$ & $11.94$ $m/s$ & $12.31$ $m/s$ & $12.47$ $m/s$ & $13.10$ $m/s$ \\
\midrule
\multirow{2}{2.4cm}{\centering Safety-Guard Intervention Ratio} & T & $\textbf{2.7\%}$ & $5.6\%$ & $5.3\%$ & $5.4\%$ & $6.2\%$ & / \\
 & A & $\textbf{2.2\%}$ & $3.3\%$ & $3.0\%$ & $2.9\%$ & $2.3\%$ & / \\
\bottomrule
\end{tabular}
\label{tab:my-table}
\end{table*}

\begin{table*}[ht!]
\caption{Collision rate statistics for different algorithms}\label{tab:Collision rate statistics for different algorithms}
\centering
\renewcommand{\arraystretch}{1.5} 
\begin{tabular}{>{\centering\arraybackslash}m{1.4cm}>{\centering\arraybackslash}m{1.2cm}>{\centering\arraybackslash}m{1.2cm}
>{\centering\arraybackslash}m{1.6cm}>{\centering\arraybackslash}m{1.2cm}>{\centering\arraybackslash}m{1.2cm}
>{\centering\arraybackslash}m{1.2cm}>{\centering\arraybackslash}m{1.4cm}}
\toprule
{\centering Stats} & Stage & \textbf{Ours} & Safety Guard & FOCOPS & SAC & CPO & TRPO-Lag\\
\midrule
\multirow{2}{*}{\centering Collision Rate} & T & $\textbf{0\%}$ & $0.12\%$ & $51.5\%$ & $67.8\%$ & $83.5\%$ & $83.4\%$ \\
 & A & $\textbf{0\%}$ & $0\%$ & $53.0\%$ & $41\%$ & $90\%$ &$83\%$\\
\bottomrule
\end{tabular}

\label{tab:my-table}
\end{table*}

\subsection{Simulation results}

The algorithm is deployed in the training environment. The computer is equipped with an Intel core i7-10700 CPU, NVIDIA GeForce GTX 1660 SYPER GPU.

Fig. \ref{Visualization_results_of_the_DD-RQ} shows the visualization results of the DD-RQ model. The experiment is carried out in a dynamic dense traffic scenario and a mixed traffic scenario including pedestrian crossing. The results of risk quantification are represented as a percentage, and the results of importance ranking are represented by different colors (red represents high importance, and green represents low importance). The dashed orange lines point to the two vehicles with the highest ranking of importance. It can be seen from Fig. \ref{Visualization_results_of_the_DD-RQ}(a) that the proposed method can output reasonable risk quantization results and reasonable importance ranking results according to different complex traffic scenarios. Fig. \ref{Visualization_results_of_the_DD-RQ}(b) shows the results of the DD-RQ model in a mixed traffic scenario. It can be seen that in the traffic scenario containing randomly appearing pedestrians, the proposed method pays attention to the pedestrians and vehicles that may generate risks in $0s$ to $8s$ and outputs lower risk quantization values in $8s$ to $24s$. This indicates that the proposed method can cope with more types of traffic participants and be useful.

In this paper, the DD-RQ model is used to combine with the safe self-evolution algorithm for autonomous driving. The benchmarks consist of various conventional safety RL algorithms, as well as a model-free RL algorithm. Among the safety RL algorithms, four mainstream methods from two major categories are selected as benchmarks \cite{liu2023datasets}, including CPO \cite{achiam2017constrained}, FOCOPS \cite{zhang2020first}, TRPO-lag \cite{stooke2020responsive}, and Safety Gard method \cite{10421882}. Meanwhile, the SAC algorithm is chosen among the model-free RL. These algorithms are utilized for decision-making and control tasks in autonomous driving. We demonstrate the effectiveness of our proposed algorithm by comparing it to these algorithms.

The average returns of our method and baseline methods are shown in Fig. \ref{average return}. It can be seen that after 100,000 training steps, the average return of the proposed algorithm steadily increased, surpassing all baseline algorithms. This is because compared with the baseline methods, our method avoids unnecessary trial-and-error collisions while ensuring the safety of the training process. Moreover, our method does not overly focus on being conservative in experience collection and policy updates, resulting in a higher overall average return.

The results of the ablation experiments are shown in Table \ref{tab:Results of the ablation experiments}, using the collision rate, the average speed and the safety guard intervention ratio as the evaluation metrics. Table \ref{tab:Results of the ablation experiments} shows the experiment results comparing the  proposed  adjustable safety limits method with fixed-parameter safety guard method and model-free RL algorithm during the training stage (T) and application stage (A). As shown in Table \ref{tab:Results of the ablation experiments}, during the training stage, the model-free RL algorithm cause a very high collision rate (49.65\%) due to no added safety mechanism, which is almost unacceptable for autonomous driving. The fixed-parameter safety guard method combines the safety mechanism, which makes the collision almost never happen (note: the aggressive parameter selection still brings the collision rate of 2\% to 3\%), but from the average speed and the safety guard intervention ratio, the policy is relatively conservative, which affects the agent efficiency. The proposed algorithm improves the average speed while ensuring no collision at all, and reduces the safety guard intervention ratio from more than 5.3\% to 2.7\%, which shows that the proposed algorithm can well suppress the policy conservative problem caused by the regular safety RL algorithm while ensuring safety. Similar conclusions can also be drawn from the experiment results of the application stage of the algorithm, that is, the proposed algorithm can ensure the improvement of the average speed of the vehicle and the decrease of the safety guard intervention ratio in the application process, which indicates that the performance of the safety RL algorithm is guaranteed.

Table \ref{tab:Collision rate statistics for different algorithms} shows collision rate statistics for different algorithms. The result shows the comparison between the proposed method and the baseline methods for scenario b) (including the complex mixed traffic scenario with pedestrian crossing). It can be seen that in challenging autonomous driving scenarios, the baseline algorithms cannot guarantee safety during training and application, and even after the algorithms converge, they still fail to accomplish the task. In contrast, the proposed method in this paper can guarantee safety, and such effect enhancement is more obvious in complex scenarios.

The proposed algorithm is deployed in the simulation environment, and spatio-temporal diagrams depicting the trajectories of the ego vehicle and traffic vehicles under scenario a) and scenario b) are presented in Fig. \ref{Spatio-temporal diagram_sa} and Fig. \ref{Spatio-temporal diagram_sb}, respectively. In these diagrams, traffic vehicles and the ego vehicle are represented as rectangular boxes of varying colors, plotted at 0.1-second intervals. As shown in Fig. \ref{Spatio-temporal diagram_sa}, the proposed algorithm effectively controls the ego vehicle, showcasing reasonable acceleration, deceleration, lane changes, and overtaking maneuvers while considering interactions within the traffic environment, demonstrating the method's effectiveness. Fig. \ref{Spatio-temporal diagram_sb} further presents the algorithm's capability to handle mixed traffic scenarios, including random pedestrian crossings, by adjusting steering and throttle to safely maneuver around pedestrians and promptly respond to traffic flow changes. This highlights the algorithm's ability to handle more types of traffic participants and cope well with challenging traffic scenarios.


\subsection{Real Vehicle Test Results}

In order to further illustrate the effectiveness of the proposed algorithm, the real vehicle test is carried out. The algorithm is deployed on an autonomous vehicle equipped with rich sensors and high-performance computers, as shown in Fig. \ref{Real_vehicle_test_Results} (a). The hardware system of the vehicle includes integrated navigation system, Lidar, millimeter-wave radar, camera, power supply, industrial computer, etc. The CPU of the industrial computer is Intel i9-9900k and the GPU is RTX-3080Ti, including Ethernet and CAN interfaces. The software architecture is based on ROS, a distributed software framework for robot development. The experimental site is selected in the closed test site of Tongji University in Shanghai, which includes 430m standardized test road sections, which can meet the test requirements of the algorithm.

As shown in Fig. \ref{Real_vehicle_test_Results} (b), we use high-fidelity simulation software and real autonomous vehicle to build this self-evolution platform based on virtual-real interaction, and realize the preliminary verification of the self-evolution algorithm on this platform. Specifically, autonomous vehicles run in both the real and virtual world, dealing with both real and virtual traffic scenarios. This scheme can ensure the security of the testing process and can improve the verification efficiency of the algorithm through the flexible configuration of the simulation environment.

The experiment results are shown in Fig. \ref{Real_vehicle_test_Results} (c)(d)(e). Fig.\ref{Real_vehicle_test_Results} (b) and (c) show the results of the real vehicle test in the real world and the virtual world, and Fig.\ref{Real_vehicle_test_Results} (e) shows the sensor recording results. It can be seen that the proposed algorithm can control the vehicle to run synchronously in the real world and the virtual world, and at the same time can realize safe and reasonable lane changing overtaking action.

\section{Conclusion}

This paper proposed a safe self-evolution algorithm for autonomous driving based on a data-driven risk quantification (DD-RQ) model. The method realized the safety situation estimation of the environment through a data-driven approach, and combined a safety-evolution decision-control integration algorithm with adjustable safety limits, which enabled the agent to ensure safety exploration without sacrificing the performance improvement ability. The proposed method was verified in a dense three-lane lane-changing scene. Experiments show that compared with the model-free RL algorithm and the regular safety RL algorithm, the proposed method can improve the average speed  while ensuring the system collision-free in the training and verification process, and reduce the safety guard intervention ratio by 50\%. Real vehicle test results show that the proposed algorithm can control the real autonomous vehicle to output safe, smooth and reasonable lane-changing overtaking actions. In the future research, we will combine large-scale natural driving data sets to extract the key features of the risk quantification model, and consider introducing the receding horizon optimization technology to improve the optimality of the algorithm.

\section*{Appendix}

\subsection{Quintic and quartic polynomial curves}
Quintic and quartic polynomial curves are respectively applied to describe the process of longitudinal and lateral trajectory planning:

\begin{equation}
\label{cal_polynomials 1}
\left[ {\begin{array}{*{20}{c}}
1&0&0&0&0&0\\
0&1&0&0&0&0\\
0&0&2&0&0&0\\
1&t&{{t^2}}&{{t^3}}&{{t^4}}&{{t^5}}\\
0&1&{2t}&{3{t^2}}&{4{t^3}}&{5{t^4}}\\
0&0&2&{6t}&{12{t^2}}&{20{t^3}}
\end{array}} \right] \cdot {p_d} = \left[ {\begin{array}{*{20}{c}}
{{d_0}}\\
{{{\dot d}_0}}\\
{{{\ddot d}_0}}\\
{{d_{fn}}}\\
{{{\dot d}_{fn}}}\\
{{{\ddot d}_{fn}}}
\end{array}} \right],\
\end{equation}

\begin{equation}
\label{cal_polynomials 2}
\left[ {\begin{array}{*{20}{c}}
1&0&0&0&0\\
0&1&0&0&0\\
0&0&2&0&0\\
0&1&{2t}&{3{t^2}}&{4{t^3}}\\
0&0&2&{6t}&{12{t^2}}
\end{array}} \right] \cdot {p_s} = \left[ {\begin{array}{*{20}{c}}
{{s_0}}\\
{{{\dot s}_0}}\\
{{{\ddot s}_0}}\\
{{{\dot s}_f}}\\
{{{\ddot s}_f}}
\end{array}} \right],\
\end{equation}

\begin{equation}
{p_d} = {\left[ {\begin{array}{*{20}{c}}
{a_d^0}&{a_d^1}&{a_d^2}&{a_d^3}&{a_d^4}&{a_d^5}
\end{array}} \right]^T},\
\end{equation}

\begin{equation}
{p_s} = {\left[ {\begin{array}{*{20}{c}}
{a_s^0}&{a_s^1}&{a_s^2}&{a_s^3}&{a_s^4}
\end{array}} \right]^T},\
\end{equation}
where $t$ is time variable, $p_d$ is coefficients of quintic polynomials for lateral planning, $[ {\begin{array}{*{20}{c}}
{{d_0}}&{{{\dot d}_0}}&{{{\ddot d}_0}}&{{d_{fn}}}&{{{\dot d}_{fn}}}&{{{\ddot d}_{fn}}}
\end{array}} ]^T$ is the boundary condition of a quintic polynomial, $p_s$ is coefficients of quartic polynomials for longitudinal planning and $[{\begin{array}{*{20}{c}}
{{s_0}}&{{{\dot s}_0}}&{{{\ddot s}_0}}&{{{\dot s}_f}}&{{{\ddot s}_f}}
\end{array}}]^T$ is the boundary condition of a quartic polynomial.

\subsection{Reward function Design}
The speed reward is designed to guide the autonomous vehicle to travel within a desired speed interval. Considering the whole process of the vehicle starting from 0 to the appropriate speed, this paper designs the speed reward into two phases, namely, the starting phase and the maintaining phase. The speed interval of the starting phase is defined as ${v_x} \in \left[ {0,{v_{s\min }}} \right]$, and the speed interval of the maintenance phase is ${v_x} \in \left[ {{v_{s\min }},{v_{s\max }}} \right]$, where ${v_{s\min }}$ and $v_{s\max }$ are the lower and upper speed limit of the maintaining phase.

For efficiency considerations, the reward function is designed as follows: the higher the speed, the higher the reward. The speed reward function is designed as:

\begin{equation}
{r_s} = {\rho _{{s_p}}} \cdot {v_{s_p}} + {\rho _{m_p}} \cdot {v_{m_p}},\
\end{equation}
where ${\rho _{{s_p}}}$ is the reward factor when the vehicle is in the starting phase, ${\rho _{{m_p}}}$ is the reward factor when the vehicle is in the maintenance phase, $v_{s_p}$ is the normalized vehicle speed in the starting phase and ${v_{m_p}}$ is the normalized vehicle speed in the maintenance phase.

Collision-free is the primary requirement for autonomous vehicles. Therefore, the agent should be penalized when it collides with surrounding objects. The collision penalty ${r_c}$ is designed as follows:

\begin{equation}
{r_c} = \left\{ {\begin{array}{*{20}{c}}
0&{\begin{array}{*{20}{c}}
{not}&{collision}
\end{array}}\\
{{\rho _{coll}}}&{collision}
\end{array}}, \right.\
\end{equation}
where ${\rho _{coll}}$ is collision penalty factor.

Considering that autonomous vehicles should ideally travel along the centerline of the lane, the lane deviation penalty ${r_{ld}}$  is designed as follows:

\begin{equation}
{r_{ld}} = {\rho _{ld}} \cdot {d_{ld}},\
\end{equation}
where ${\rho _{ld}}$ is the lane deviation penalty factor and $d_{ld}$ is the relative distance between the ego vehicle and the nearest lane. That is, the farther the ego vehicle is from the lane, the greater the penalty.

\begin{table}[!t]
\caption{Reward Coefficients Settings for ${\pi _\phi }$\label{tab:Reward Coefficients Settings}}
\centering
\begin{tabular}{cc}
\toprule
Reward coefficients terms & Symbol \& Value \\
\midrule
Starting phase & ${\rho _{{s_p}}} = 0.5$ \\
Maintaining phase  & ${\rho _{m_p}} = 4.0$ \\
Collision  & ${\rho _{coll}} =  - 50.0$ \\
Lane Deviation  & ${\rho _{ld}} =  - 1.0$ \\
\bottomrule
\end{tabular}
\label{Reward_Coeff}
\end{table}

All reward coefficients s for ${\pi _\phi }$ are listed in Table \ref{Reward_Coeff}.

\subsection{Experiment Setting}

The hyperparameters for the simulation is shown in Table \ref{tab:Hyperparameters for the simulation}.

\begin{table}[!t]
\caption{Hyperparameters for the simulation\label{tab:Hyperparameters for the simulation}}
\centering
\begin{tabular}{>{\centering\arraybackslash}m{3.7cm} >{\centering\arraybackslash}m{4.2cm}}

\toprule
Parameters & Symbol \& Value \\
\midrule
Simulation step size & $\Delta t = 0.1s$ \\
Discount factor of value network  & $\gamma  = 0.99$ \\
Learning rate of adjust temperature & $\lambda  = 3e - 4$ \\
Learning rate of target network & $\tau  = 5e - 3$ \\
Value network layers number & $n_{value} = 3$  \\
Policy network layers number & $n_{policy} = 4$  \\
Value network neurons number in hidden layer& $n_{vwidth} = 256$  \\
Policy network neurons number in hidden layer& $n_{pwidth} = 256$  \\
Total training steps & ${T_{total}} = 1000000$ \\
Driving agent dataset size & ${n_{DA}} = 300000$ \\
Lane width & ${L_{width}} = 3.5m$ \\
Maximum and Minimum $T_c$ & ${T_{c\max }} = 4.0s$, ${T_{c\min }} = 0.5s$ \\
Maximum and Minimum  acceleration & ${a_{\max }} = 2.0m/{s^2}, {a_{\min }} = -2.0m/{s^2}$ \\
Linearization parameter $k_1$,$b_1$  & ${k_1} =  - \frac{7}{{200}}$, ${b_1} = 4.0$ \\
Linearization parameter $k_2$,$b_2$&  ${k_2} =  0.92$, ${b_2} = 0.02$ \\
\bottomrule
\end{tabular}
\end{table}

\section*{Acknowledgments}
We would like to thank the National Key R\&D Program of China under Grant No2022YFB2502900, National Natural Science Foundation of China (Grant Number: U23B2061), the Fundamental Research Funds for the Central Universities of China, Xiaomi Young Talent Program, and thanks the reviewers for the valuable suggestions

\bibliographystyle{IEEEtran}
\bibliography{document}

\vspace{-10 mm}
\begin{IEEEbiography}[{\includegraphics[width=1in,height=1.25in,clip,keepaspectratio]{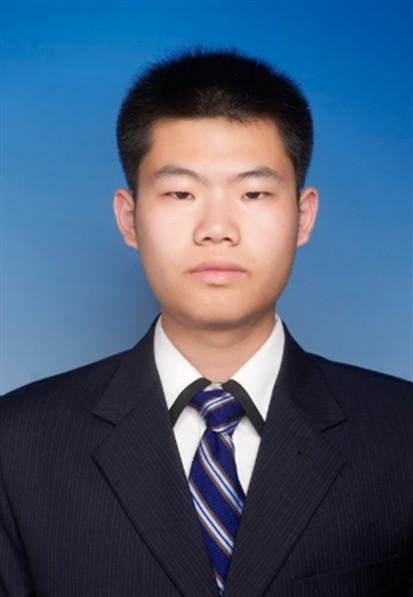}}]{Shuo Yang}
received the B.S. and M.S degree (cum laude) from the College of Automotive Engineering, Jilin University, Changchun, China, in 2017. He is currently pursuing the Ph.D. degree in School of Automotive Studies, Tongji University, Shanghai. His research interests include reinforcement learning,  autonomous vehicle, intelligent transportation system and vehicle dynamics.
\end{IEEEbiography}

\vspace{-10 mm}

\begin{IEEEbiography}[{\includegraphics[width=1in,height=1.25in,clip,keepaspectratio]{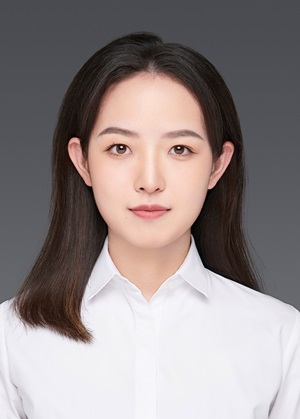}}]{Shizhen Li}
Shizhen Li received the B.S. degree (cum laude) from the School of Automotive Studies, Tongji University, Shanghai, China, in 2022. She is currently pursuing the M.S. degree in School of Automotive Studies, Tongji University, Shanghai, China. Her research interests include reinforcement learning and autonomous vehicle and intelligent transportation system.
\end{IEEEbiography}

\vspace{ -10  mm}

\begin{IEEEbiography}[{\includegraphics[width=1in,height=1.25in,clip,keepaspectratio]{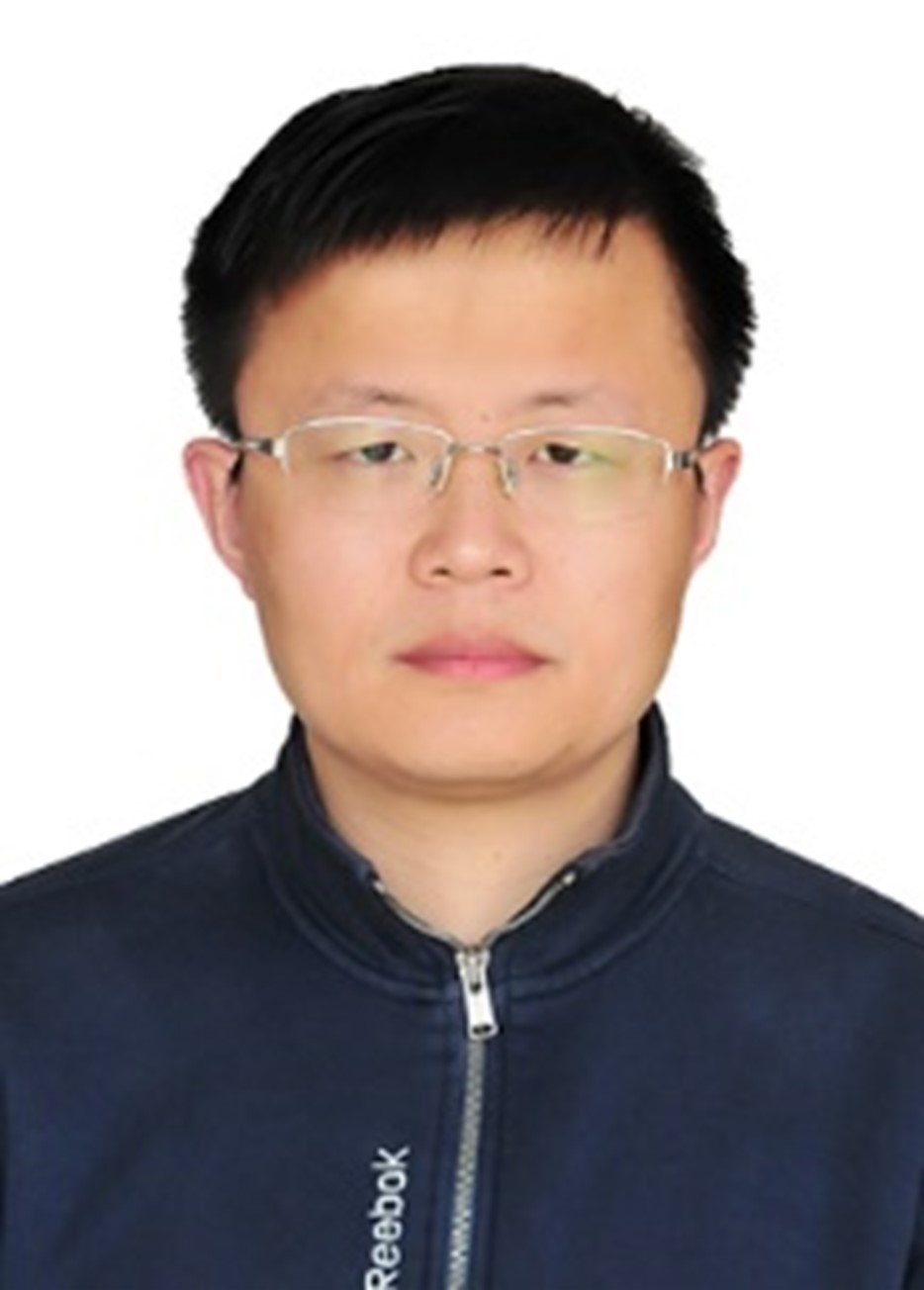}}]{Yanjun Huang}
is a Professor at School of Automotive studies, Tongji University. He received his PhD Degree in 2016 from the Department of Mechanical and Mechatronics Engineering at University of Waterloo. His research interest is mainly on autonomous driving and artificial intelligence in terms of decision-making and planning, motion control, human-machine cooperative driving.

He has published several books, over 80 papers in journals and conference; He is the recipient of IEEE Vehicular Technology Society 2019 Best Land Transportation Paper Award. He is serving as AE of IEEE/TITS, P I MECH ENG D-JAUT, IET/ITS, SAE/IJCV, Springer Book series of connected and autonomous vehicle, etc.
\end{IEEEbiography}

\vspace{ -10mm}

\begin{IEEEbiography}[{\includegraphics[width=1in,height=1.25in,clip,keepaspectratio]{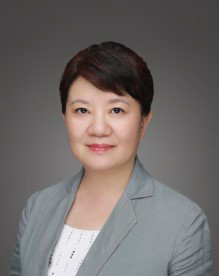}}]{Hong Chen }
(M'02-SM'12-F'22) received the B.S. and M.S. degrees in process control from Zhejiang University, China, in 1983 and 1986, respectively, and the Ph.D. degree in system dynamics and control engineering from the University of Stuttgart, Germany, in 1997. In 1986, she joined Jilin University of Technology, China. From 1993 to 1997, she was a Wissenschaftlicher Mitarbeiter with the Institut fuer Systemdynamik und Regelungstechnik, University of Stuttgart. Since 1999, she has been a professor at Jilin University and hereafter a Tang Aoqing professor. Recently, she joined Tongji University as a distinguished professor. Her current research interests include model predictive control, nonlinear control, artificial intelligence and applications in mechatronic systems e.g. automotive systems.
\end{IEEEbiography}
\vspace{-5 mm}

\end{document}